\documentclass[final]{cvpr}

\usepackage{times}
\usepackage{epsfig}
\usepackage{graphicx}
\usepackage{graphics}
\usepackage{amsmath}
\usepackage{amssymb}
\usepackage{lipsum}
\usepackage{booktabs}
\usepackage{pifont}%
\usepackage[dvipsnames]{xcolor}
\usepackage{bbm}
\usepackage{subcaption}
\usepackage{bbm}
\usepackage{multirow}

\usepackage[pagebackref=true,breaklinks=true,colorlinks,bookmarks=false]{hyperref}

\def\cvprPaperID{3268} %
\def\confYear{CVPR 2021}

\newcommand{\arka}[1]{{}}

\newcommand{\tkfull}[0]{Visual Semantic Role Labeling in Videos}
\newcommand{\tk}[0]{VidSRL}
\newcommand{\dsn}[0]{VidSitu}

\newcommand{\cmark}{\ding{51}}%
\newcommand{\xmark}{\ding{55}}%

\newcommand{\ar}[1]{\textit{\mbox{#1}}}

\newcommand{\agb}[1]{[#1]}
\newcommand{\best}[1]{\textbf{\underline{#1}}}
\newcommand{\sbest}[1]{\textbf{#1}}

\newcommand{\vweb}[0]{\href{https://vidsitu.org/}{\texttt{vidsitu.org}}}

\begin{document}

\title{Visual Semantic Role Labeling for Video Understanding}

\author{
Arka Sadhu$^{1 \dagger}$ \quad \quad Tanmay Gupta$^3$ \quad \quad Mark Yatskar$^{2}$ \quad \quad Ram Nevatia$^1$ \quad \quad Aniruddha  Kembhavi$^3$\\
$^1$University of Southern California \quad \quad $^2$University of Pennsylvania \quad \quad $^3$PRIOR @ Allen Institute for AI\\

{\tt\small {\{asadhu|nevatia\}@usc.edu} \quad myatskar@seas.upenn.edu \quad \{tanmayg|anik\}@allenai.org}
}

\twocolumn[{
\renewcommand\twocolumn[1][]{#1}
\maketitle
\vspace*{-0.9cm}
\begin{center}
% \href{https://vidsitu.org/}{\texttt{vidsitu.org}}
\vweb{}\\    
\end{center}
\includegraphics[width=\linewidth]{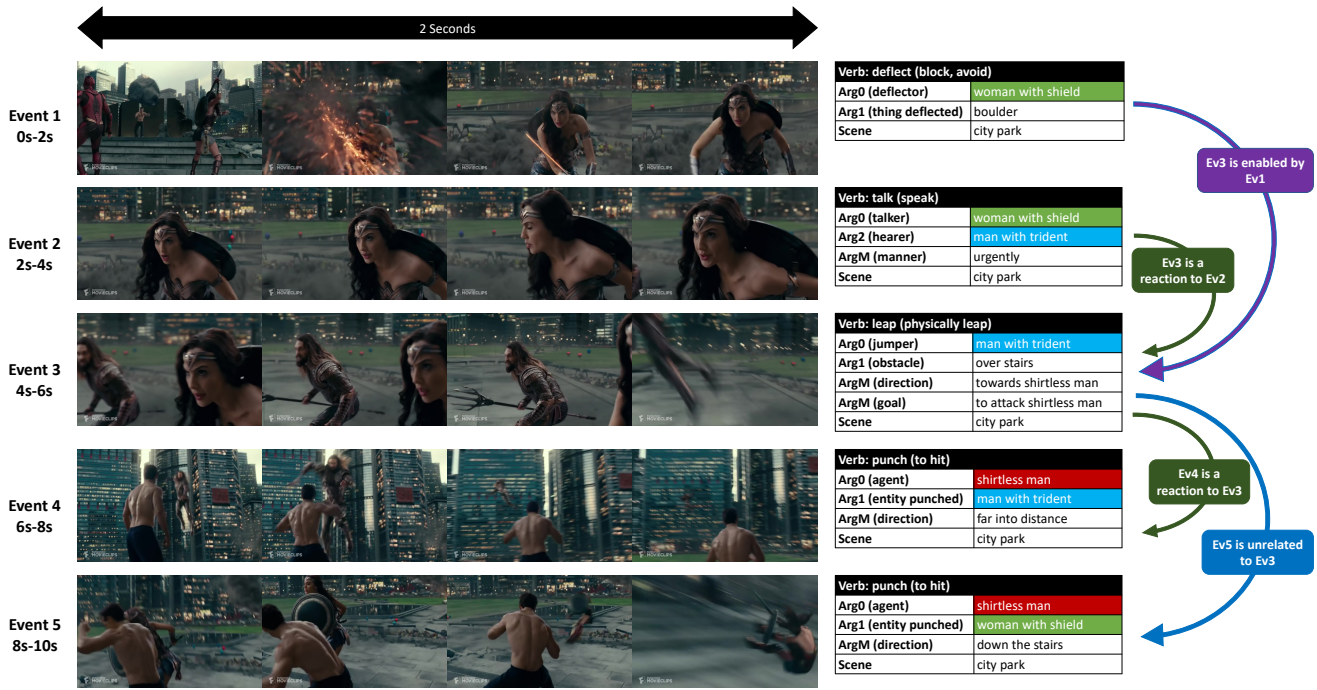}
\captionof{figure}{\textbf{A sample video and annotation from \dsn{}.} The figure shows a 10-second video annotated with 5 events, one for each 2-second interval. 
Each event consists of a verb (like ``deflect'') and its arguments (like \ar{Arg0 (deflector)} and \ar{Arg1 (thing deflected)}). 
Entities that participate in multiple events within a clip are co-referenced across all such events (marked using the same color).
Finally, we relate all events to the central event (Event 3).
The video can be viewed at: \url{https://youtu.be/3sP7UMxhGYw?t=20}
(from 20s-30s).
} 
\label{fig:intro_fig}

\vspace*{0.5cm}
}]

\renewcommand*{\thefootnote}{$\dagger$}
\footnotetext{Part of the work was done during Arka's internship at PRIOR@AI2}
\renewcommand*{\thefootnote}{\arabic{footnote}}

\maketitle
\begin{abstract}
\label{s:abstract}

We propose a new framework for understanding and representing related salient events in a video using visual semantic role labeling. 
We represent videos as a set of related events, wherein each event consists of a verb and multiple entities that fulfill various roles relevant to that event. 
To study the challenging task of semantic role labeling in videos or \tk{}, we introduce the \dsn{} benchmark, a large scale video understanding data source with $29K$ $10$-second movie clips richly annotated with a verb and semantic-roles every $2$ seconds. 
Entities are co-referenced across events within a movie clip and events are connected to each other via event-event relations. 
Clips in \dsn{} are drawn from a large collection of movies (${\sim}3K$) and have been chosen to be both complex (${\sim}4.2$ unique verbs within a video) as well as diverse (${\sim}200$ verbs have more than $100$ annotations each). 
We provide a comprehensive analysis of the dataset in comparison to other publicly available video understanding benchmarks,  several illustrative baselines and evaluate a range of standard video recognition models. 
Our code and dataset is available at \vweb{}.

\end{abstract}
\vspace{-1em}
\section{Introduction}
\label{s:intro}

Videos record events in our lives with both short and long temporal horizons. These recordings frequently relate multiple events separated geographically and temporally and capture a wide variety of situations involving human beings interacting with other humans, objects and their environment.
Extracting such rich and complex information from videos can drive numerous downstream applications such as describing videos \cite{krishna2017dense,Xu2016MSRVTTAL,Wang2019VaTeXAL}, answering queries about them \cite{yu2019activityqa,xu2017videoqa}, retrieving visual content \cite{miech19howto100m}, 
building knowledge graphs~\cite{Mahon2020KnowledgeGE} and 
even teaching embodied agents to act and interact with the real world~\cite{Young2020VisualIM}.

Parsing video content is an active area of research with much of the focus centered around 
tasks such as action classification \cite{Kay2017TheKineticsH}, localization \cite{Heilbron2015ActivityNetAL} and spatio-temporal detection \cite{Gu2018AVAAV}.
Although parsing human actions is a critical component of understanding videos, actions by themselves paint an incomplete picture, missing critical pieces such as the agent performing the action, the object being acted upon, the tool or instrument used to perform the action, location where the action is performed and more. %
Expository tasks such as video captioning and story-telling provide a more holistic understanding of the visual content; but akin to their counterparts in the image domain, they lack a clear definition of the type of information being extracted making them notoriously hard to evaluate~\cite{kilickaya-etal-2017-evaluating,Vinyals2017ShowAT}.%

Recent work in the image domain \cite{yatskar2016,Pratt2020GroundedSR,Gupta2015VisualSR} has attempted to move beyond action classification via the task of visual semantic role labeling - producing not just the primary activity in an image or region, but also the entities participating in that activity via different roles. 
Building upon this line of research, we propose \tk{} -- the task of recognizing spatio-temporal situations in video content.
As illustrated in Figure.~\ref{fig:intro_fig}, \tk{} involves recognizing and temporally localizing salient events across the video, identifying participating actors, objects, and locations involved within these events, co-referencing these entities across events over the duration of the video, and relating how events affect each other over time. 
We posit that \tk{}, a considerably more detailed and involved task than action classification with more precise definitions of the extracted information than video captioning, is a step towards obtaining a holistic understanding of complex videos. 

To study \tk{}, we present \dsn{}, a large video understanding dataset of over $29$K videos drawn from a diverse set of $3$K movies. 
Videos in \dsn{} are exactly $10$ seconds long and are annotated with $5$ verbs, corresponding to the most salient event taking place within the five 2 second intervals in the video. 
Each verb annotation is accompanied with a set of roles whose values
\footnote{Following nomenclature introduced in ImSitu\cite{yatskar2016}, every verb (deflect) has a set of roles (Arg0 deflector, Arg1 thing deflected) which are realized by noun values. Here, we use ``value'' to refer to free-form text used describing the roles (woman with shield, boulder).}
are annotated using free form text. 
In contrast to verb annotations which are derived from a fixed vocabulary, the free form role annotations allow the use of referring expressions (\emph{e.g. boy wearing a blue jacket}) to disambiguate entities in the video. An entity that occurs in any of the five clips within a video is consistently referred to using the same expression, allowing us to develop and evaluate models with co-referencing capability. 
Finally, the dataset also contains event relation annotations capturing causation (Event Y is Caused By/Reaction To Event X) and contingency (Event X is a pre-condition for Event Y).
The key highlights of \dsn{} include: (i) \emph{Diverse Situations:}  \dsn{} enjoys a large vocabulary of verbs (1500 unique verbs curated from PropBank~\cite{Palmer2005PropBank} with $200$ verbs having at least $100$ event annotations) and entities ($5600$ unique nouns with $350$ nouns occurring in at least $100$ videos); 
(ii) \emph{Complex Situations:} Each video is annotated with $5$ inter-related events and has an average of $4.2$ unique verbs, $6.5$ unique entities and; 
(iii) \emph{Rich Annotations:} \dsn{} provides structured event representations ($3.8$ roles per event) with entity co-referencing and event-relation labels.

To facilitate further research on \tk{}, we provide a comprehensive benchmark that supports partwise evaluation of various capabilities required for solving \tk{} and create baselines for each capability using state-of-art architectural components to serve as a point of reference for future work. 
We also carefully choose metrics that provide a meaningful signal of progress towards achieving competency on each capability. 
Finally, we perform a human-agreement analysis that reveals a significant room for improvement on the \dsn{} benchmark.

Our main contributions are:
(i) the \tk{} task formalism for understanding complex situations in videos;
(ii) curating the richly annotated \dsn{} dataset that consists of diverse and complex situations for studying \tk{};
(iii) establishing an evaluation methodology for assessing crucial capabilities needed for \tk{} and establishing baselines for each using state-of-art components. The dataset and code are publicly available at \vweb{}.

\begin{table*}[t!]
\centering
\resizebox{\linewidth}{!}{%
\begin{tabular}{lll}
\toprule
\textbf{Task}             & \textbf{Required Annotations}                                                 & \textbf{Dataset}                               \\
\midrule
Action Classification     & Action Labels                                                                 & Kinetics\cite{Kay2017TheKineticsH}, 
ActivityNet \cite{Heilbron2015ActivityNetAL},
Moments in Time \cite{Monfort2020MomentsIT},
Something-Something\cite{Goyal2017TheS}, HVU \cite{diba2019holistic}
\\
Action Localization       & Action Labels, Temp. Segments                                              & ActivityNet, Thumos\cite{Idrees2017TheTC},
HACS \cite{zhao2019hacs},
Tacos\cite{Regneri2013GroundingADTacos}, Charades\cite{Sigurdsson2016HollywoodIH}, COIN\cite{Tang2019COINAL}     \\
Spatio-Temporal Detection & Action Labels, Temp. Segments, BBoxes                              & AVA\cite{Gu2018AVAAV}, AVA-Kinetics\cite{Li2020TheAL}, EPIC-Kitchens \cite{Damen2018ScalingEV}, JHMDB\cite{Jhuang:ICCV:2013}                            \\
Video Description         & Captions, Temp. Segments                                                   & ActivityNet\cite{Heilbron2015ActivityNetAL}, Vatex\cite{Wang2019VaTeXAL}, YouCook\cite{DaXuDoCVPR2013}, MSR-VTT \cite{Xu2016MSRVTTAL} , LSMDC \cite{Rohrbach2015ADF}         \\
Video QA                  & Q/A, Subtitle or Script (optional)                            & MSRVTT-QA\cite{xu2017videoqa}, VideoQA\cite{Zeng2017LeveragingVD}, ActivityNetQA\cite{yu2019activityqa}, TVQA\cite{Lei2018TVQALC}, MovieQA\cite{Tapaswi2016MovieQAUS} \\
Text to Video Retrieval   & Text Query, ASR output (optional)                                             & HowTo100M\cite{miech19howto100m}, TVR\cite{Lei2020TVRAL}, DiDeMo\cite{Hendricks2017LocalizingMI}, Charades-STA\cite{Gao2017TALLTA}                              \\
Video Object Grounding    & Text Query, Temp. Segments, BBoxes                              & ActivityNet-SRL\cite{Sadhu2020VideoOG}, YouCookII\cite{Zhou2018WeaklySupervisedVO}, VidSTG \cite{Zhang2020WhereDI},VID-sentence\cite{Chen2019WeaklySupervisedSG}       \\
\midrule
VidSRL                    & Verbs, SRLs, Corefs, Event Relations, Temp. Segments & VidSitu                                       \\
\bottomrule
\end{tabular}%
}
\vspace{-0.5em}
\caption{A non-exhaustive summary of video understanding tasks, required annotations and benchmarks.
}
\vspace{-1em}
\label{tab:vid_unds_cmp}

\end{table*}

\section{Related Work}
\label{s:rel_work}

\textbf{Video Understanding}, a fundamental goal of computer vision, is an incredibly active area of research involving a wide variety of tasks such as action classification~\cite{Carreira2017QuoVA,Feichtenhofer2019SlowFastNF, Wang2016TemporalSN}, localization~\cite{BSN2018arXiv,Lin2019BMNBN} and spatio-temporal detection~\cite{Girdhar2019VideoAT}, video description~\cite{Wang2019VaTeXAL,krishna2017dense}, question answering~\cite{yu2019activityqa}, and object grounding~\cite{Sadhu2020VideoOG}. 
Tasks like detecting atomic actions at $1$ second intervals~\cite{Girdhar2019VideoAT,Wu2019LongTermFB,Sun2018ActorCentricRN} are short horizon tasks whereas ones like summarizing $180$ second long videos~\cite{Zhou2018EndtoEndDV} are extremely long horizon tasks. 
In contrast, our proposed task of \tk{} operates on $10$ second video at $2$ second intervals. It entails producing a verb for the salient activity within each $2$ second interval as well as predicting multiple entities that fulfill various roles related to that event, and finally relating these events across time.

In support of these tasks, the community has also proposed datasets \cite{Kay2017TheKineticsH, Heilbron2015ActivityNetAL,Gu2018AVAAV}, over the past few years. 
While early datasets were small datasets with several hundred or thousand examples\cite{Soomro2012UCF101AD,Kuehne2011HMDBAL}, recent datasets are massive\cite{miech19howto100m} enabling researchers to train large neural models and also employ pre-training strategies\cite{miech19endtoend,Zhu2020ActBERTLG,Li2020HEROHE}. Section~\ref{s:dataset}, Table~\ref{tab:ds_comp} and Figure~\ref{fig:dstats_cmp} provide a comparison of our proposed dataset to several relevant datasets in the field.
Due to space constraints, we are unable to provide a thorough description of all the relevant work. Instead we point the reader to relevant surveys on video understanding~\cite{aafaq2019video,Kong2018HumanAR,Zhang2019ACS} and also present a holistic overview of tasks and datasets in Table~\ref{tab:vid_unds_cmp}.

\textbf{Visual Semantic Role Labeling} has been primarily explored in the image domain under situation recognition~\cite{yatskar2016,Pratt2020GroundedSR}, visual semantic role labeling~\cite{Gupta2015VisualSR,Li2020CrossmediaSC,Silberer2018GroundingSR} and human-object interaction~\cite{Chao2015HICOAB,Chao2018LearningTD}.
Compared to images, visual semantic role labeling in videos requires not just recognizing actions and arguments at a single time step but
aggregating information about interacting entities across frames, co-referencing the entities participating across events.

\textbf{Movies for Video Understanding:}
The movie domain serves as a rich data source for spatio-temporal detection \cite{Gu2018AVAAV}, movie description \cite{Rohrbach2015ADF}, movie question answering \cite{Tapaswi2016MovieQAUS} , story-based retrieval \cite{bain2020condensed}, generating social graphs \cite{moviegraphs} tasks, and classifying shot style \cite{huang2020movienet}. 
In contrast to a lot of this prior work, we focus only on the visual activity of the various actors and objects in the scene, \ie no additional modalities like movie-scripts, subtitles or audio are presented in our dataset.

\section{\tk{}: The Task}
\label{s:task}
State-of-the-art video analysis capabilities like video activity recognition and object detection yield a fairly impoverished understanding of videos by reducing complex events involving interactions of multiple actors, objects, and locations to a bag of activity and object labels. While video captioning promises rich descriptions of videos, the open-ended task definition of captioning lends itself poorly to a systematic representation of such events and evaluation thereof. The motivation behind \tk{} is to expand the video analysis toolbox with vision models that produce richer yet structured representations of complex events in videos than currently possible through video activity recognition, object detection, or captioning.

\textbf{Formal task definition.} Given a video $V$, \tk{} requires a model to predict a set of related salient events $\{E_i\}_{i=1}^k$ constituting a situation. Each event $E_i$ consists of a verb $v_i$ chosen from a set of of verbs $\mathcal{V}$ and values (entities, location, or other details pertaining to the event described in text) assigned to various roles relevant to the verb. We denote the roles or arguments of a verb $v$ as $\{A_j^v\}_{j=1}^m$ and $A_j^v{\leftarrow}a$ implies that the $j^{th}$ role of verb $v$ is assigned the value $a$. In Fig.~\ref{fig:intro_fig} for instance, event $E_1$ consists of verb $v{=}\text{``deflect (block, avoid)"}$ with \ar{Arg0 (deflector)} $\leftarrow$ ``woman with shield". The roles for the verbs are obtained from PropBank~\cite{Palmer2005PropBank}. Finally, we denote the relationship between any two events $E$ and $E'$ by \mbox{$l(E,E')\in \mathcal{L}$} where $\mathcal{L}$ is an event-relations label set. We now discuss simplifying assumptions and trade-offs in designing the task.

\textbf{Timescale of Salient Events.} 
What constitutes a salient event in a video is often ambiguous and subjective. For instance given the 10 sec clip in Fig.~\ref{fig:intro_fig}, one could define fine-grained events around atomic actions such as  ``turning" (Event 2 third frame) or take a more holistic view of the sequence as involving a ``fight''. This ambiguity due to lack of constraints on timescales of events makes annotation and evaluation challenging. We resolve this ambiguity by restricting the choice of salient events to one event per fixed time-interval. Previous work on recognizing atomic actions~\cite{Gu2018AVAAV} relied upon 1 sec intervals. 
An appropriate choice of time interval for annotating events is one that enables rich descriptions of complex videos while avoiding incidental atomic actions. 
We observed qualitatively that a 2 sec interval strikes a good balance between obtaining descriptive events and the objectiveness needed for a systematic evaluation. Therefore, for each 10 sec clip, we annotate 5 events $\{E_i\}_{i=1}^5$.
Appendix \ref{ss:app_data_curate} elaborates on this choice.

\textbf{Describing an Event.} We describe an event through a verb and its arguments. For verbs, we follow recent work in action recognition like ActivityNet~\cite{Heilbron2015ActivityNetAL} and Moments in Time~\cite{Monfort2020MomentsIT} that choose a verb label for each video segment from a curated list of verbs. To allow for description of a wide variety of events, we select a large vocabulary of $2.2K$ visual verb from PropBank~\cite{Palmer2005PropBank}. Verbs in PropBank are diverse, distinguish between homonyms using verb-senses (e.g. ``strike (hit)'' vs ``strike (a pose)''), and provide a set of roles for each verb.
We allow values of arguments for the verb to be free-form text. This allows disambiguation between different entities in the scene using referring expression such as ``man with trident" or ``shirtless man" (Fig.~\ref{fig:intro_fig}). Understanding of a video may require consolidating partial information across multiple views or shots. In \tk{}, while the 2 sec clip is sufficient to assign the verb, roles may require information from the whole video since some entities involved in the event may be occluded or lie outside the camera-view for those 2 secs but are visible before or after. For e.g., in Fig~\ref{fig:intro_fig} Event 2, information about ``Arg2 (hearer)'' is available only in Event 3.

\textbf{Co-Referencing Entities Across Events.}
Within a video, an entity may be involved in more than one event, for instance, ``woman with shield'' is involved in Events 1, 2, and 5 and ``man with trident'' is involved in Events 2, 3, and 4. In such cases, we expect \tk{} models to understand co-referencing \ie a model must be able to recognize that the entity participating across those events is the same even though the entity may be playing different roles in those events. Ideally, evaluating coreferencing capability requires grounding entities in the video (e.g. using bounding boxes). Since grounding entities in videos is an expensive process, we currently require the phrases referring to the same entity across multiple events within each 10 sec clip to match exactly for coreference assessment. See supp. for details on how coreference is enforced in our annotation pipeline.

\textbf{Event Relations.} Understanding a video requires not only recognizing individual events but also how events affect one another. Since event relations in videos is not yet well explored, we propose a taxonomy of event relations as a first step -- inspired by prior work on a schema for event relations in natural language \cite{Hong2016BuildingAC} that includes ``Causation'' and ``Contingency''. In particular, if Event B follows (occurs after) Event A, we have the following relations: 
(i) \textit{Event B is caused by Event A} (Event B is a direct result of Event B);
(ii) \textit{Event B is enabled by Event A} (Event A does not cause Event B, but Event B would not occur in the absence of Event A);
(iii) \textit{Event B is a reaction to Event A} (Event B is a response to Event A); and
(iv) \textit{Event B is unrelated to Event A}
(examples are provided in supplementary).

\section{\dsn{} Dataset}
\label{s:dataset}
To study \tk{}, we introduce the \dsn{} dataset that offers videos with \textbf{diverse} and \textbf{complex}
situations (a collection of related events) and \textbf{rich} annotations with verbs, semantic roles, entity co-references, and event relations. 
We describe our dataset curation decisions (Section \ref{sec:data_prep}) followed by analysis of the dataset (Section \ref{sec:data_stats}).

\subsection{Dataset Curation}~\label{sec:data_prep}
\label{ss:dat_prep}
We briefly describe the main steps in the data curation process and provide more information in Appendix \ref{s:app_data_coll}.

\textbf{Video Source Selection.} 
Videos from movies are well suited for \tk{} since they are naturally diverse (wide-range of movie genres) and often involve multiple interacting entities.
Also, scenarios in movies typically play out over multiple shots which makes movies a challenging testbed for video understanding.
We use videos from Condensed-Movies \cite{bain2020condensed} which collates videos from MovieClips- a licensed YouTube channel containing engaging movie scenes.

\textbf{Video Selection.}
Within the roughly 1000 hours of MovieClips videos, we select 30K diverse and interesting 10sec videos to annotate while avoiding visually uneventful segments common in movies such as actors merely engaged in dialogue. This selection is performed using a combination of human detection, object detection and atomic action prediction followed by a sampling of no more than 3 videos per movieclip after discarding inappropriate content.

\textbf{Curating Verb Senses.}
We begin with the entire PropBank \cite{Palmer2005PropBank} vocabulary of ${\sim}6k$ verb-senses.
We manually remove fine-grained and non-visual verb-senses and further discard verbs that do not appear in the MPII-Movie Description (MP2D) dataset~\cite{Rohrbach2015ADF} (verbs extracted using a semantic-role parser~\cite{Shi2019SimpleBM}). This gives us a set of $2154$ verb-senses.

\textbf{Curating Argument Roles.}
We wish to establish a set of argument roles for each verb-sense. We initialize the argument list for each verb-sense using Arg0, Arg1, Arg2 arguments provided by PropBank and then expand this using frequently used (automatically extracted) arguments present in descriptions provided by the MP2D dataset.

\begin{table}[t]
\centering
\scriptsize
\resizebox{\linewidth}{!}{
\begin{tabular}{@{}c|ccccc|c@{}}
\toprule
                      & Train  & Valid & Test-Vb & Test-SRL & Test -ER & Total  \\ \midrule
\# Movies             & 2431   & 151   & 151       & 153      & 151        & 3037   \\
\# Videos             & 23626  & 1326  & 1353      & 1598     & 1317       & 29220  \\
\# Clips              & 118130 & 6630  & 6765      & 7990     & 6585       & 146100 \\
\# Verbs Ann / Clip   & 1      & 10    & 10        & 10       & 10         &        \\
\# Verb Ann           & 118130 & 66300 & 67650     & 79900    & 65850      & 397830 \\
\# Unique Verb Tuples & 23196  & 1317  & 1341      & 1571     & 1299       & 28724  \\
\# Values Ann / Role  & 1      & 3     & 3         & 3        & 3          &        \\
\# Role Ann           & 118130 & 19890 & 20295     & 23970    & 19755      & 202040 \\ \bottomrule
\end{tabular}
}
\caption{\textbf{Statistics} on splits of \dsn{}. Note that \dsn{} contains multiple verb and role annotations for val and test sets for accurate evaluation. 
}
\vspace{-1.5em}
\label{tab:vsitu_ds_stats}
\end{table}
\begin{table*}[t]
\centering
\resizebox{\linewidth}{!}{%
\begin{tabular}{@{}lcccccccccccc@{}}
\toprule
Dataset & Domain & SRLs, Coref       & EvRel           & Videos & Clips  & Descr. & Descr./Clip (Train) & Avg. Clip Len. (s) & Uniq Vbs/Vid  & Uniq Ents/Vid & Avg. Roles/Event \\ \midrule
MSR-VTT     & open  & Implicit & \xmark & 7k    & 10k   & 200k & 20 & 14.83 & 1.88 & 2.80 & 1.56 \\
MPII-MD     & movie & Implicit & \xmark & 94    & 68k   & 68.3 & 1  & 3.90  & 1.87 & 2.99 & 2.24 \\
ActyNet-Cap & open  & Implicit & \xmark & 20k   & 100k  & 100k & 1  & 36.20 & 2.30 & 3.75 & 2.37 \\
Vatex-en    & open  & Implicit & \xmark & 41.3k & 41.3k & 413k & 10 & 10.00 & 2.69 & 4.04 & 1.96 \\
\dsn{} & movie  & \textbf{Explicit} & \textbf{\cmark} & 29.2k  & 146k & 146k & 1           & 10.00          & \textbf{4.21} & \textbf{6.58} & \textbf{3.83}    \\ \bottomrule

\end{tabular}
}
\caption{\textbf{Dataset statistics across video description datasets.} We \textbf{highlight} key differences from previous datasets such as explicit SRL, co-reference, and event-relation annotations, and greater diversity and density of verbs, entities, and semantic roles. For a fair comparison, for all datasets we use a single description per video segment when more than one are available. 
}
\label{tab:ds_comp}
\end{table*}

\begin{figure*}[!ht]
    \centering
    \includegraphics[width=0.9\linewidth]{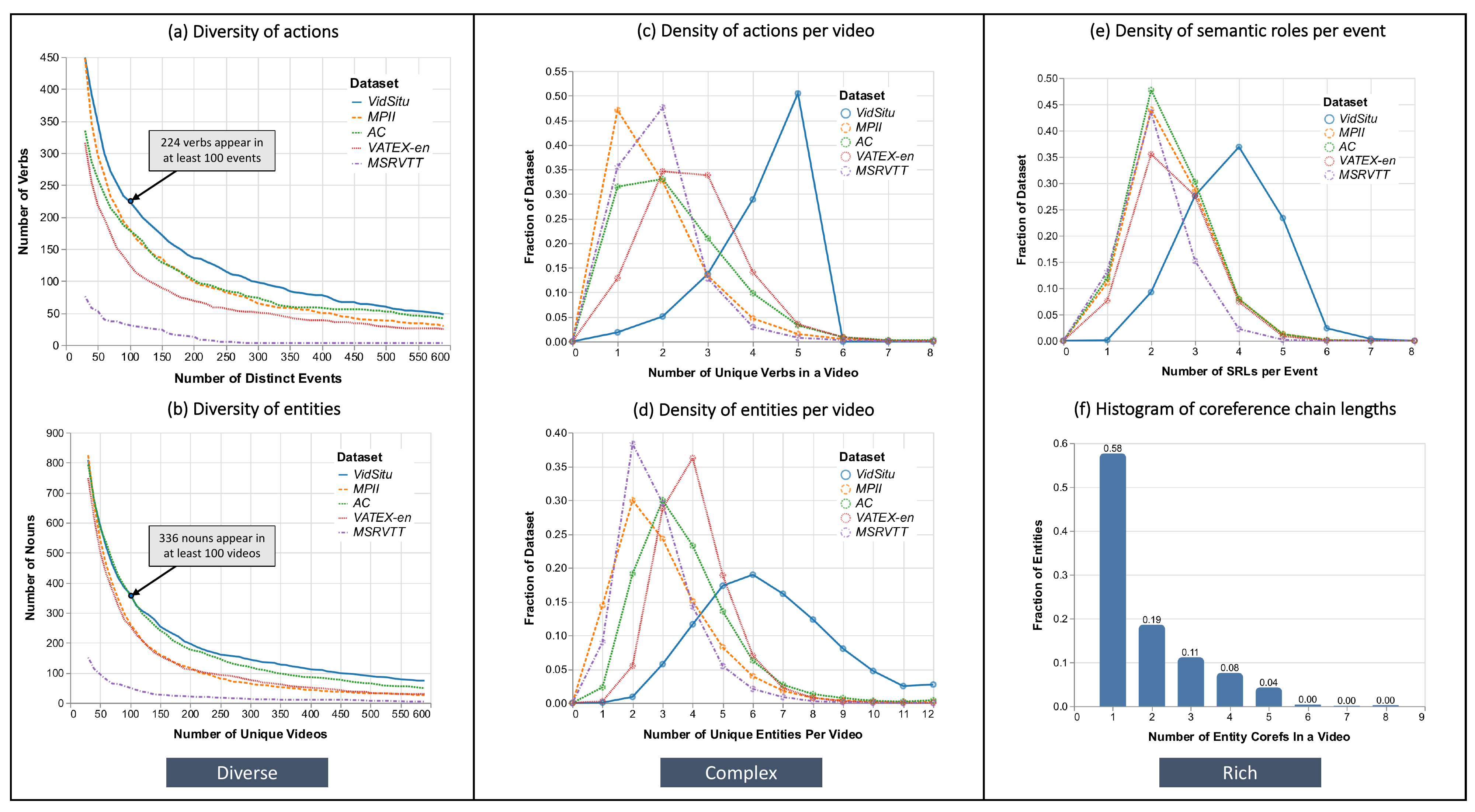}
    \caption{\textbf{Data analysis.} An analysis of \dsn{} in comparison to other large scale relevant video datasets. We focus on the \textbf{diversity} of actions and entities in the dataset (a and b), the \textbf{complexity} of the situations measured in terms of the number of unique verbs and entities per video (c and d) and the \textbf{richness} of annotations (e and f).} 
    \label{fig:dstats_cmp}
    \vspace{-1em}
\end{figure*}

\textbf{Annotations.}
Annotations for the verbs, roles and relations are obtained via Amazon Mechanical Turk (AMT). The annotation interface enables efficient annotations while encouraging rich descriptions of entities and enabling a reuse of entities through the video (to preserve co-referencing).
See Appendix \ref{ss:app_ann_pipe} for details.

\textbf{Dataset splits.} 
\dsn{} is split into train, validation and test sets via a $80{:}5{:}15$ split, ensuring that videos from the same movie end up in exactly one of those sets. 
Table \ref{tab:vsitu_ds_stats} summarizes these statistics of these splits.
We emphasize that each of the three tasks namely \textbf{Verb Prediction}, \textbf{Semantic Role Prediction and Co-Referencing} and \textbf{Event Relation Prediction} have separate test sets.

\textbf{Multiple Annotations for Evaluation Sets.} Via controlled trials (see Sec~\ref{ss:eval_metrics}) we measured the annotation disagreement rate for the train set.
Based on this data, we obtain multiple annotations for validation and test sets using a 2-stage annotation process. In the first stage, we collect 10 verbs for each 2 second clip (1 verb per worker). In the second stage, we get role labels for the verb with the highest agreement from 3 different workers. %

\subsection{Dataset Analysis and Statistics}~\label{sec:data_stats}
\vspace{-1em}

We present an extensive analysis of \dsn{} focusing on three key elements: (i) \textbf{diversity} of events represented in the dataset; (ii) \textbf{complexity} of the situations; and (iii) \textbf{richness} of annotations. We provide comparisons to four prominent video datasets containing text descriptions -- MSR-VTT~\cite{Xu2016MSRVTTAL}, MPII-Movie Description~\cite{Rohrbach2015ADF}, ActivityNet Captions~\cite{krishna2017dense}, and Vatex-en~\cite{Wang2019VaTeXAL} (the subset of descriptions in English). 
Table~\ref{tab:ds_comp} summarizes basic statistics from all datasets. 
For consistency, we use one description per video segment whenever multiple annotations are available, as is the case for Vatex-en, MSR-VTT, validation set of ActivityNet-Captions and both validation and test sets of \dsn{}. For datasets without explicit verb or semantic role labels, we extract these using a semantic role parser~\cite{Shi2019SimpleBM}.

\textbf{Diversity of Events.} To assess the diversity of events represented in the dataset, we consider cumulative distributions of verbs\footnote{As a fair comparison to datasets which do not have senses associated with verbs, we collapse verb senses into a single unit for this analysis.} and nouns (see Fig.~\ref{fig:dstats_cmp}-a,b). For any point $n$ on the horizontal axis, the curves show the number of verbs or nouns with at least $n$ annotations. \dsn{} not only offers greater diversity in verbs and nouns as compared to other datasets but also a large number of verbs and nouns occur sufficiently frequently to enable learning useful representations. For instance, $224$ verbs and $336$ nouns have at least 100 annotations. In general, since movies inherently intend to engage viewers, movie datasets such as MPII and \dsn{} are more diverse than open-domain datasets like ActivityNet-Captions and VATEX-en.

\textbf{Complexity of Situations.} We refer to a situation as complex if it consists of inter-related events with multiple entities fulfilling different roles across those events. To evaluate complexity, Figs.~\ref{fig:dstats_cmp}-c,d compare the number of unique verbs and entities per video across datasets. Approximately, $80\%$ of videos in \dsn{} have at least 4 unique verbs and $70\%$ have 6 or more unique entities, in comparison to $20\%$ and $30\%$ respectively for VATEX-en. Further, Fig.~\ref{fig:dstats_cmp}-e shows that $90\%$ of events in \dsn{} have at least 4 semantic roles in comparison to only $55\%$ in VATEX-en. Thus, situations in \dsn{} are considerably more complex that existing datasets.

\textbf{Richness of Annotations.} While existing video description datasets only have unstructured text descriptions, \dsn{} is annotated with rich structured representations of events that includes verbs, semantic role labels, entity coreferences, and event relations. Such rich annotations not only allow for more thorough evaluation of video analysis techniques but also enable researchers to study relatively unexplored problems in video understanding such as entity coreference and relational understanding of events in videos. Fig.~\ref{fig:dstats_cmp}-f shows the fraction of entity coreference chains of various lengths.

\begin{figure}
    \centering
    \includegraphics[width=0.9\columnwidth]{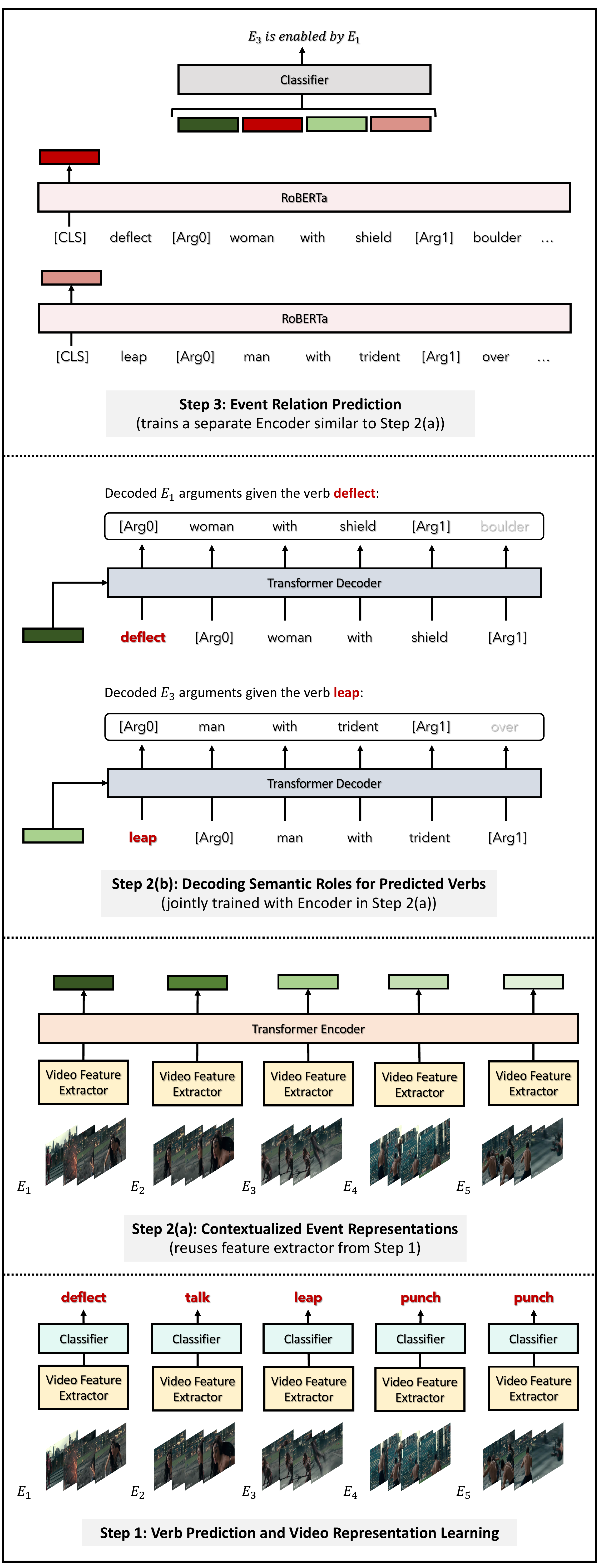}
    \caption{\textbf{Models.} The figure illustrates our baselines for verb, semantic role, and event prediction using state-of-the-art network components such as SlowFast~\cite{Feichtenhofer2019SlowFastNF} network for video feature extraction, transformers~\cite{Vaswani2017AttentionIATransformer} for encoding events in a video and verb-conditional decoding of roles, and RoBERTa~\cite{Liu2019RoBERTaAR} language encoder for event-relation prediction.}
    \label{fig:mdl_fig}
\end{figure}

\section{Baselines}
\label{s:method}
For a given video, \tk{} requires predicting verbs and semantic roles for each event as well as event relations. We provide powerful baselines to serve as a point of comparison these crucial capabilities. These models leverage architectures from state-of-the-art video recognition models.

\textbf{Verb Prediction.}
Given a 2 sec clip, we require a model to predict the verb corresponding to the most salient event in the clip. As baselines, we provide state-of-art action recognition models such as I3D \cite{Carreira2017QuoVA} and SlowFast \cite{Feichtenhofer2019SlowFastNF} networks (Step 1 in Fig.~\ref{fig:mdl_fig}). We consider variants of I3D both with and without Non-Local blocks~\cite{Wang2018NonlocalNN} and for SlowFast networks, we consider variants with and without the Fast channel. For each architecture, we train a model from scratch as well as a model finetuned after pretraining on Kinetics~\cite{Kay2017TheKineticsH}. All models are trained with a cross-entropy loss over the set of action labels. For subsequent stages, these verb classification models are frozen and used as feature extractors.

\textbf{Argument Prediction Given Verbs:}
Given a 10 sec video and a verb for each of the 5 events, a model is required to infer entities and their roles involved in each event.  
To this end, we adapt seq-to-seq models \cite{Sutskever2014SequenceTS} that consist of an encoder and a decoder (Step 2(a,b) in Fig.~\ref{fig:mdl_fig}). 
Specifically, independent event features are fed through a transformer~\cite{Vaswani2017AttentionIATransformer} encoder (TxEnc) to get contextualized event representations. 
Then for each event, the corresponding encoded representation and the verb are passed to a transformer decoder (TxDec) to generate the sequence of arguments and roles for that event. 
As an example, for Event 1 in Fig~\ref{fig:intro_fig}, we expect to generate the following sequence: { {\small\agb{Arg0} woman with shield \agb{Arg1} boulder \agb{Scene} city park}}

The generated sequence is post-processed to obtain the argument role structure similar to those of the annotations Figure~\ref{fig:intro_fig}. We also provide language only baselines using our TxDec architecture as well as a GPT2 decoder.

\textbf{Event Relation Prediction:}
A model must infer how the various events within a video are related given the verb and arguments. For a pair of ordered events $(E_i,E_j)$ with $i<j$, with corresponding verbs and semantic roles, we construct a multimodal representation of each event denoted by $m_i$ and $m_j$ (Step 3 in Fig.~\ref{fig:mdl_fig}). Each of these representations is a concatenation of visual representation from TxEnc and a language representation of the sequence of verbs, arguments, and roles obtained from a pretrained RoBERTa~\cite{Liu2019RoBERTaAR}-base language model. $m_i$ and $m_j$ are concatenated and fed through a classifier to predict the event relation.

\vspace{-0.7em}
\section{Experiments}
\label{s:exp}
\dsn{} allows us to evaluate performance in 3 stages: (i) verb prediction; (ii) prediction of semantic roles with coreferencing given the video and verbs for each event; and (iii) event relations prediction given the video and verbs and semantic roles for a pair of events.

\begin{table*}[ht]
\centering
\scriptsize
\resizebox{\linewidth}{!}{
\begin{tabular}{@{}c|cc|cccccc|cccccc@{}}
\toprule
\multirow{2}{*}{Model} & \multirow{2}{*}{Vis} & \multirow{2}{*}{Enc} & \multicolumn{6}{c|}{Val}                          & \multicolumn{6}{c}{Test}                         \\
                       &                      &                      & C     & R-L   & C-Vb  & C-Arg & Lea   & Lea-S & C     & R-L   & C-Vb  & C-Arg & Lea   & Lea-S \\
                       \midrule
GPT2                   & \xmark               & \xmark               & 34.67 & 40.08 & 42.97 & 34.45 & 48.08 & 28.1  & 36.48 & 41.33 & 44.27 & 36.51 & 49.38 & 30.24 \\
TxDec                  & \xmark               & \xmark               & 35.68 & 41.19 & 47.5  & 32.15 & \best{51.76} & 28.6  & 35.34 & 41.45 & 44.44 & 32.06 & \best{52.46} & 29.18 \\
Vid TxDec              & SlowFast             & \xmark               & 44.78 & 40.61 & 49.97 & 41.24 & 37.88 & 28.69 & 44.95 & 41.12 & 49.46 & 41.98 & 38.91 & 30.21 \\
Vid TxEncDec           & SlowFast             & \cmark               & 45.52 & \best{42.66} & \best{55.47} & \best{42.82} & \sbest{50.48} & \sbest{31.99} & 47.25 & \best{43.46} & \sbest{52.92} & \best{45.48} & \sbest{50.88} & \sbest{33.5}  \\
Vid TxDec              & I3D                  & \xmark               & \best{47.14} & 40.67 & 51.61 & 41.29 & 37.89 & 30.38 & \sbest{47.9}  & 41.5  & 51.29 & 43.62 & 38.77 & 31.73 \\
Vid TxEncDec           & I3D                  & \cmark               & \sbest{47.06} & \sbest{42.41} & \sbest{51.67} & \sbest{42.76} & 48.92 & \best{33.58} & \best{48.51} & \sbest{42.96} & \best{53.88} & \sbest{44.53} & 49.61 & \best{35.46} \\
\midrule
Human*                 &                      &                      & 84.85 & 39.77 & 91.7  & 80.15 & 72.1  & 70.33 & 83.68 & 40.04 & 87.78 & 79.29 & 71.77 & 70.6 \\
\bottomrule
\end{tabular}
}
\caption{\textbf{Semantic role prediction and co-referencing metrics.} Vis. denotes the visual features used (\xmark $\;$ if not used), and Enc. denotes if video features are contextualized.
C: CIDEr, R-L: ROUGE-L, C-Vb: CIDEr scores averaged across verbs, C-Arg: CIDEr scores averaged over arguments.
Lea-S: Lea-soft. See Section ~\ref{ss:eval_metrics} for details.
}
\label{tab:vbarg_fullds}
\end{table*}

\begin{table}[ht]
\centering
\resizebox{\linewidth}{!}{
\begin{tabular}{@{}lc|ccc|ccc@{}}
\toprule
\multirow{2}{*}{Model} & \multirow{2}{*}{Kin.} & \multicolumn{3}{c|}{Val}      & \multicolumn{3}{c}{Test}     \\ 
            &       & Acc@1 & Acc@5 & Rec@5 & Acc@1 & Acc@5 & Rec@5 \\ \midrule
I3D                    & \xmark      & 31.18  & 67.00  & 5.24  & 31.91  & 67.36  & 5.33   \\
I3D+NL                 & \xmark      & 30.17  & 66.83  & 4.88  & 31.43  & 67.70  & 5.02   \\
Slow+NL                & \xmark      & 33.05  & 68.83  & 5.82  & 34.29  & 69.56  & 6.24   \\
SlowFast+NL            & \xmark      & 32.64  & 69.22  & 6.11  & 33.94  & \sbest{70.54}  & 6.56   \\
\midrule
I3D                    & \cmark      & 29.65  & 60.77  & 18.21 & 29.87  & 59.10  & 19.54  \\
I3D+NL                 & \cmark      & \sbest{39.40}  & \sbest{70.82}  & 17.12 & \sbest{38.42}  & 69.27  & 18.46  \\
Slow+NL                & \cmark      & 29.05  & 58.69  & \sbest{19.19} & 29.03  & 58.77  & \sbest{21.06}  \\
SlowFast+NL            & \cmark      & \best{46.79}  & \best{75.90}  & \best{23.38} & \best{46.37}  & \best{75.28}  & \best{25.78}  \\ \bottomrule
\end{tabular}
}
\caption{\textbf{Verb classification metrics.} Acc@K: Event Accuracy considering $10$ ground-truths and $K$ model predictions. Rec@K: Macro-Averaged Verb Recall with K predictions. Kin. denotes whether Kinetics is used.}
\label{tab:vb_full_ds}
\vspace{-1em}
\end{table}

\begin{table}[ht]
\centering
\resizebox{\linewidth}{!}{
\begin{tabular}{@{}ccccc@{}}
\toprule
          & Verb   & Args   & Val Macro-Acc & Test Macro-Acc \\ \midrule

Roberta   & \cmark & \cmark & 25.00                            & 25.00                             \\
TxEnc     & \cmark & \cmark & 25.00                            & 25.00                             \\
\midrule
Vid TxEnc & \xmark & \xmark & 34.13         & 30.97          \\
Vid TxEnc & \xmark & \cmark & \best{34.54}  & \sbest{32.89}  \\
Vid TxEnc & \cmark & \cmark & \sbest{34.15} & \best{32.98}  \\
\bottomrule
\end{tabular}
}
\caption{\textbf{Event relation classification metrics.} Macro-Averaged Accuracy on Validation and Test Sets. We evaluate only on the subset of data where two annotators agree.}
\label{tab:evrel_cmp}
\vspace{-1em}
\end{table}

\subsection{Evaluation Metrics}
\label{ss:eval_metrics}

In \tk{}, multiple outputs are plausible for the same input video. This is because of inherent ambiguity in the choice of verb used to describe the event (e.g. the same event may be described by ``fight", ``punch" or ``hit"), and the referring expression used to refer to entities in the video (e.g. ``boy with black hair" or ``boy in the red shirt"). We confirm this ambiguity through a human-agreement analysis on a subset of 100 videos (500 events) with 25 verb annotations and 5 role annotations per event. Importantly, through careful manual inspection we confirm that a majority of differences in annotation for the same video across AMT workers are due to this inherent ambiguity and not due to a lack of annotation quality.

\textbf{Verb Prediction.} The ambiguity in verbs associated with events suggests that commonly used metrics such as Accuracy, Precision, and F1 are ill suited for the verb prediction task as they would penalize correct predictions that may not be represented in the ground truth annotations. However, recall based metrics such as Recall@k are suitable for this task. Since the large verb vocabulary in \dsn{} presents a class-imbalance challenge, we use a macro-averaged Recall@k that better reflects performance across all verb-senses instead of focusing on dominant classes.  %

We now describe our macro-averaged Verb Recall@k metric. For any event, we only consider the set of verbs which appears at least twice within the ground-truth annotations (each event in val and test sets has $10$ verb annotations). 
For event $E_j$ (where $j$ indexes events in our evaluation set), let this set of agreed-upon ground-truth be denoted by $G_j$. 
We compute recall@k for each verb-sense $v_i\in \mathcal{V}$ (where $i$ indexes verb-senses in the vocabulary $\mathcal{V}$) as 
\begin{align}
    R_i^k &= \frac{\sum_j \mathbbm{1}(v_i \in G_j)\times \mathbbm{1}(v_i \in P_j^k)}{\sum_j \mathbbm{1}(v_i \in G_j)}
\end{align}
where $\mathbbm{1}$ is an indicator function and $P_j^k$ denotes the set of top-k verb predictions for $E_j$. Macro-averaged verb recall@k is given by $\frac{1}{|\mathcal{V}|}\sum_i R_i^k$. 
We report macro-average verb recall@5 (R@5) but also report top-1 and top-5 accuracy (Acc@1/5) for completeness.

\textbf{Semantic Role Prediction and Co-referencing.} Given a video and verb for each event, we wish to measure the semantic role prediction performance. Through a human-agreement analysis we discard arguments such as direction (ADir) and manner (AMnr) which do not have a high inter-annotator agreement and retain Arg0, Arg1, Arg2, ALoc, and AScn for evaluation. This agreement computation is computed using the CIDEr metric by treating one of the chosen annotations as a hypothesis and remaining annotations as references for each argument. In addition to reporting a micro-averaged CIDEr score (C), we also compute macro-averaged CIDEr where the macro-averaging is performed across verb-senses (C-Vb) or argument-types (C-Arg). %
ROUGE-L (R-L)~\cite{lin-2004-rouge} is shown for completeness. 

Since \dsn{} provides entity coreference links across events and roles, we use LEA~\cite{Moosavi2016WhichCE} a link-based co-reference metric to measure coreferencing capability. Other metrics (MUC~\cite{Vilain1995AMC}, BCUBE~\cite{Bagga1998EntityBasedCC}, CEAFE~\cite{luo-2005-coreference}) can be found in the supp.
Co-referencing in our case is done via exact string matching over the predicted set of arguments. Thus, even if the predictions are incorrect, but just the coreference is correct, LEA would give it a higher score.
To address this, we propose a soft version of LEA termed LEA-soft (denoted with Lea-S) which assigns weights to cluster matches using their CIDEr score (defined in the supp.).

\textbf{Event-Relation Prediction Accuracy.}
Event-relation prediction is a 4-way classification problem. 
For the subset of 100 videos, We found event relations conditioned on the verbs to have $60\%$ agreement. 
For evaluation, we use the subset of event pairs for which 2 out of 3 workers agreed on the relation. We use top-1 accuracy (Acc@1) averaged across the classes as the metric for relation prediction.

\subsection{Results}
\label{ss:vb_res}

\textbf{Verb Classification:}
We report macro-averaged Rec@5 (preferred metric; Sec.~\ref{ss:eval_metrics}) and Acc@1/5 on both validation and test sets in Tab.~\ref{tab:vb_full_ds}. %
We observe verb prediction in \dsn{} follows similar trends as other action recognition tasks. Specifically, SlowFast architectures outperform I3D and Kinetics pretraining significantly and consistently improves recall across all models by $\approx10$ to $16$ points.%

\textbf{Argument Prediction:}
We report micro and macro-averaged version of CIDEr and ROUGE-L in Tab.~\ref{tab:vbarg_fullds} (see supp. for other metrics). First, video conditioned models significantly outperform video-blind baselines. Next, we observe that using an encoder to contextualize events in a video improves performance across almost all metrics. Interestingly, while SlowFast outperformed I3D in verb prediction, the reverse is true for semantic role prediction. Even so, a large gap exists between current methods and human performance.

We also evaluate coreferencing ability demonstrated by models without explicitly enforcing it during training. In Tab.~\ref{tab:vbarg_fullds}, we report both Lea and Lea-S (preferred; Sec.~\ref{ss:eval_metrics}) metrics and find that current techniques are unable to learn coreferencing directly from data. Among all models, only Vid TxEncDec outperformed a language only baseline (GPT2) on both val and test sets, leaving lots of room for improvement in future models.

\textbf{Event Relation Prediction} results are provided in Table~\ref{tab:evrel_cmp}.
Crucially, we find video-blind baselines don't train at all and end up predicting the most frequent class ``Enabled By'' (hence it gets $0.25$ for always predicting majority class).
This suggests there exists no exploitable biases within the dataset and underscores the importance and challenge posed by event relations. In contrast, video encoder models even when given just the video without any verb description outperform video-blind baselines. Adding context in the form of verb senses and arguments yields small gains.%

In summary, powerful baselines show promise on the three sub-tasks. However, it is clear that \dsn{} poses significant new challenges with a huge room for improvement.

\section{Conclusion}
\label{s:conc}

We introduce visual semantic role labeling in videos in which models are required to identify salient actions, participating entities and their roles within an event, co-reference entities across time, and recognize how actions affect each other.
We also present the \dsn{} dataset with diverse videos, complex situations, and rich annotations.

\section{Acknowledgement}
\label{s:ack}
{\footnotesize
We thank the Mechanical Turk workers for doing an outstanding
work in annotating the dataset -  without them VidSitu and the paper would not exist.
We are also grateful to the suggestions and feedback provided by the three anonymous reviewers.
This research was supported, in part, by the Office of Naval Research under grant \#N00014-18-1-2050.
}

\newpage
\appendix
\begin{center}
  {\large \bf Appendix \par}
\end{center}

\numberwithin{equation}{section}
\setcounter{table}{0}
\setcounter{figure}{0}
\setcounter{equation}{0}

Errata: 
In Figure 1, Event 2 Arg2 should be ``man with trident'' instead of ``main with trident''.

Appendix provides details on: 
\begin{enumerate}
    \itemsep0em
    \item A Brief Summary of Semantic Roles, and their usage in our paper.
    \item Details on Dataset Curation and Annotation Interface
    \item Additional Dataset Statistics
    \item Additional Implementation Details
    \item Details on Lea-Soft along with Tables with All Metrics
    \item DataSheet \cite{Gebru2018DatasheetsFD} for \dsn{}
    \item Qualitative Analysis of Data (this is attached as a video file in the zip folder).
\end{enumerate}

\section{Semantic Roles: A Brief Summary}
\label{s:app_srl_intro}

Semantic Role Labeling attempts to abstract out at a high-level who does what to whom \cite{Strubell2018LinguisticallyInformedSF}.
It is a popular natural language task which attempts at obtaining such structured outputs from natural language descriptions.
As such there are multiple sources to obtain semantic roles such as FrameNet \cite{Baker1998FrameNet}, PropBank \cite{Palmer2005PropBank} and VerbNet \cite{Brown2019VerbNet}.
Prior work on situation recognition in images (ImSitu) \cite{yatskar2016} have curated list of verbs (situations) from FrameNet, and action recognition dataset (Moments in Time) \cite{Monfort2020MomentsIT} have curated action vocabulary from VerbNet.
However, we qualitatively found both vocabulary to be insufficient to represent actions, and thus chose PropBank which contained action-oriented verbs.
As such, PropBank has been used for video object grounding \cite{Sadhu2020VideoOG} but not in the context of collecting semantic roles from visual data.

PropBank contains a set of numbered semantic roles for each verb ranging from Arg0 to Arg4.
Each numbered argument has a specific definition for a particular verb but some themes are similar across verbs (adapted from PropBank annotation guidelines \cite{bonial2010propbank}\footnote{\url{http://clear.colorado.edu/compsem/documents/propbank_guidelines.pdf}}).
For the verb ``throw'':
\begin{itemize}
    \itemsep0em
    \item Arg0: Agent -- object performing the action. For \eg ``person''
    \item Arg1: Patient -- object on which action is performed. For \eg ``ball''
    \item Arg2: Instrument, Benefactive, Attribute. For \eg ``towards a basket''
    \item Arg3: Starting Point
    \item Arg4: Ending Point
    \item ArgM: Modifier -- location (LOC), manner(MNR), direction (DIR), Purpose (PRP), Goal (GOL), Temopral (TMP), Adverb (ADV)
\end{itemize}

In general, we noticed that Arg3 and Arg4 were exceedingly rare for visual verbs, thus we restrict our attention to Arg0, Arg1, Arg2 for numbered arguments.
For modifier arguments, we found Location (LOC) to be universally valid for all video segments. Thus, for those verbs where LOC doesn't apply usually, we additionally add a semantic role ``Scene'' which refers to ``where'' the event takes place (such as ``living room'', ``near a lake'').
Other arguments were chosen based on their appearance in MPIID dataset, and we most commonly used Manner (which suggests ``how'' the action takes place) and Direction (details in the Section \ref{s:app_data_coll}).
For rest of the paper, we use ALoc, ADir, AMnr, and AScn to denote location, direction, manner and scene arguments respectively.
\section{Dataset Collection}
\label{s:app_data_coll}
In this section we describe details on dataset collection including curation of verbs and arguments, followed by details on annotation interface, quality control and reward structure.

\subsection{Dataset Curation}
\label{ss:app_data_curate}
We provide more details on Dataset Curation which were omitted from Section~\ref{ss:dat_prep}  of the main paper.

\textbf{Video Source Selection.} 
As suggested in the Section \ref{ss:dat_prep} we aimed at a domain with two criterion: the videos should be by themselves cover diverse situations (``climb'' verb should not just be associated with rocks or mountains, but also things like top of a car), and that the each video should contain complex situation (the video shouldn't depict someone doing the same task over extended period of time, which would lower chances of finding meaningful event relations and be repetitive in verbs and arguments over the entire video).

After a brief qualitative analysis, we found
instruction domain videos (HowTo100M \cite{miech19howto100m}, YouCookII \cite{Zhou2018WeaklySupervisedVO}, COIN \cite{Tang2019COINAL}) to have very fine-grained actions with less diversity and less complexity within small segments,
open domain sources (ActivityNet \cite{Heilbron2015ActivityNetAL}, Moments in Time \cite{Monfort2020MomentsIT}, Kinetics \cite{Kay2017TheKineticsH}, HACS\cite{Zhang2019ACS}) to be somewhat diverse but low complexity within a small segment. 
This led us to Movie domain which span multiple genres leading to appreciable diversity as well as complexity. 

We converged on using MovieClips \cite{bain2020condensed} rather than other movie sources such as MPII \cite{Rohrbach2015ADF}, since MovieClips already provide one-stage of filtering to provide interesting videos.
While using the same movies as used in AVA\cite{Gu2018AVAAV} was an option, we found that the video retention was quite low (around 20\% of the movie are removed from you-tube), and the movie contained long contiguous segments with low complexity. 
We also note some other datasets like MovieNet \cite{huang2020movienet}, Movie Synopsis Dataset \cite{Xiong_2019_ICCV}, Movie Graphs \cite{moviegraphs} do not provide movie videos and cannot be used for collecting annotations.
One demerit of using movie domain is that the verb distributions are skewed towards actions like ``talk'', ``walk'', ``stare''. 
Despite this we find the videos to be reasonably complex.
\begin{figure}
    \centering
    \includegraphics[width=\linewidth]{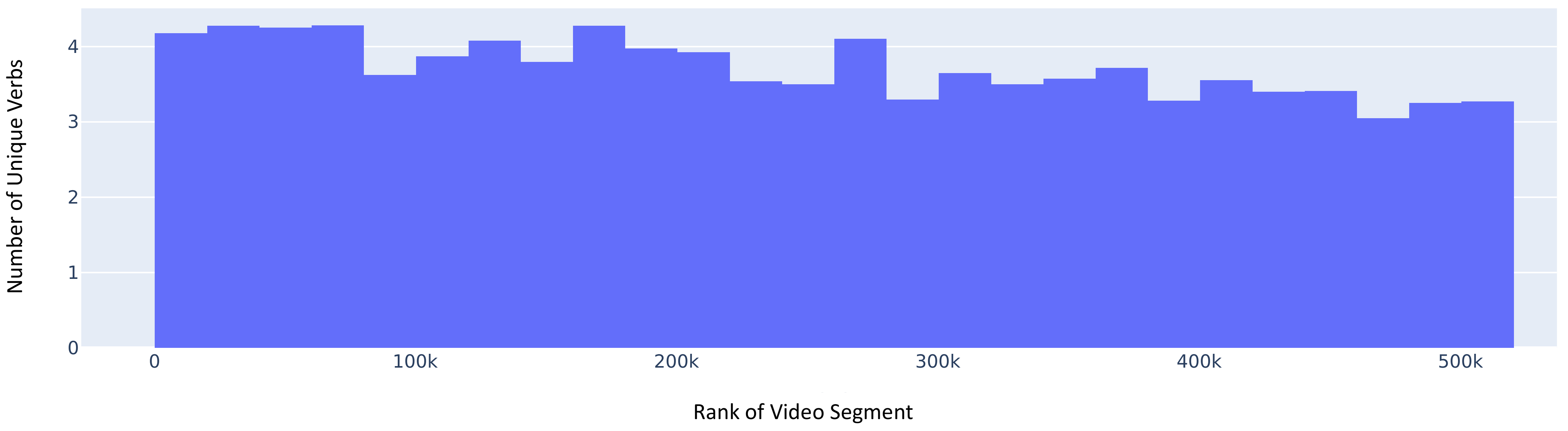}
    \caption{Bar graph showing number of unique verbs with respect to the rank of the video segment as computed via our heuristic based on predicted labels from SlowFast Network \cite{Feichtenhofer2019SlowFastNF} trained on AVA\cite{Gu2018AVAAV}. }
    \label{fig:stratified_sampling_res_stats}
\end{figure}

\begin{figure*}[t]
    \centering
    \includegraphics[width=\linewidth]{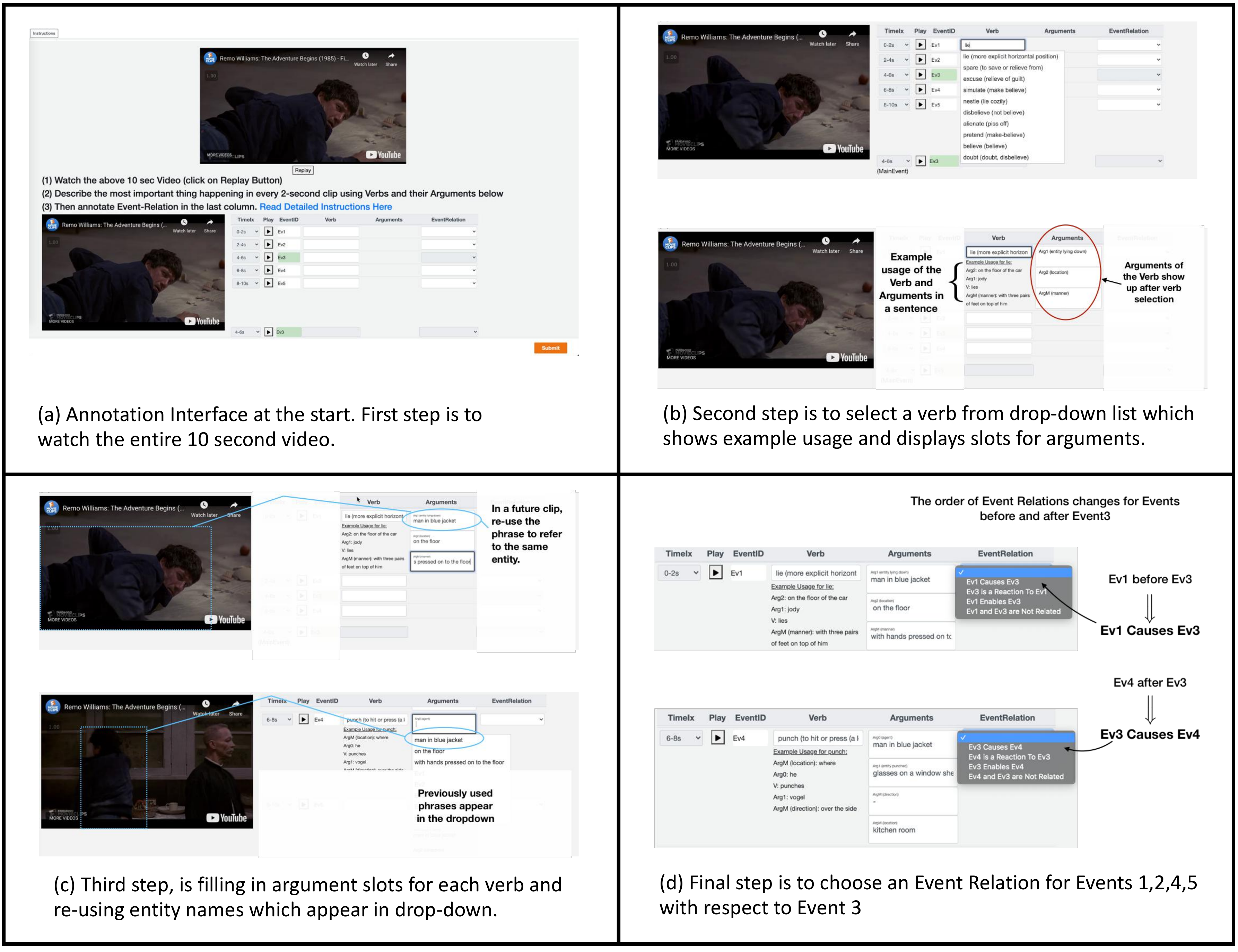}
    \caption{Illustration of our annotation interface. (a) depicts the initial screen an annotator sees. In the first step, one needs to watch the entire $10$ second video.
    (b) depicts the second step of choosing a verb from a drop-down which contains verb senses obtained from PropBank. After selecting a verb, an example usage is shown along with corresponding argument roles which need to be filled.
    (c) depicts filling the argument slots for each verb which can be phrases of arbitrary length. Each filled in phrase can be re-used in a subsequent slot, to enforce co-reference of the entities. 
    (d) shows the final step of choosing event relations once all the arguments for all events are filled. The event relations should be classified based on causality and contingency for Events 1,2,4,5 with respect to Event 3.}
    \label{fig:ann_interface}
\end{figure*}

\textbf{Video Selection.}
MovieClips spans a total of $1k$ Hours which is far beyond what can be reasonably annotated. 
To best utilize available annotation budget, we are primarily interested in identifying video segments depicting complex situations with a high precision while avoiding visually uneventful segments common in movies such as those simply involving actors engaged in dialogue.

To avoid such segments, we use the following heuristic: a video with more atomic actions per person is likely to be more eventful. 
So, we divide all movieclips into 10 second videos with a stride of 5 seconds, obtain human bounding boxes from the MaskRCNN~\cite{He2020MaskR} object detector trained on the MSCOCO~\cite{Lin2014MicrosoftCoCo} dataset, predict atomic actions for each detected person using the SlowFast~\cite{Feichtenhofer2019SlowFastNF} activity recognition model trained on the AVA~\cite{Gu2018AVAAV} dataset, and rank all videos by the average number of unique atomic actions per person in the video. 
In particular, we discard labels such as ``talk'', ``listen'', ``stand'' and ``sit'' as these atomic actions didn't correlate with complexity of situations.
Since ``action'' sequences like ``fight scenes'' are favored by our ranking measure, we use simple heuristic of removing ``martial arts'' actions to avoid oversampling such scenes  and improve diversity of situations represented in the selected videos.

To confirm the usefulness of the proposed heuristic, we conduct an experiment where we annotate $1k$ videos chosen uniformly sampled across the entire dataset (as shown in Figure \ref{fig:stratified_sampling_res_stats}).
Reducing number of unique verbs shows the effectiveness of our heuristic and suggests at least $80K$ videos segments (which translates to $27K$ non-overlapping video segments) can be richly annotated.

For final video selection, we randomly choose set of videos from the top-K ranks, such that the newly chosen videos don't overlap with already chosen videos, and that no more than $3$ videos are uploaded from the same Youtube video within a particular batch.

\textbf{Curating Verb Senses.}
To curate verb senses, we follow a two-step process: from the initial list of $\sim6k$ verb senses in PropBank \cite{Palmer2005PropBank}, first we manually filter verb senses which share the same lemmatized verb (as previously stated ``go'' has 23 verb senses) to retain only ``visual'' verb senses (for instance we remove the verb sense of ``run''  which refers to running a business).
We keep all $3.7K$ verbs with a single verb sense and of the remaining $2364$ verbs-senses (shared across $809$ verbs) we retain $629$ verb senses (shared across $561$ verbs).
Second, to further restrict the set of verbs to those useful for describing movies, we discard verbs that do no appear at all in the MPII-Movie Description (MP2D) dataset~\cite{Rohrbach2015ADF}. To extract verbs from the descriptions we use a semantic-role parser~\cite{Shi2019SimpleBM}. This results in a final set of $2154$ verb-senses.

\textbf{Curating Argument Roles.}
Once we have curated the verb-senses from PropBank, we aim to delegate a set of argument roles for each verb-sense which would be filled based on the video.
While PropBank provides numbered arguments for each verb-sense there are two issues with directly using them: first, some arguments are less relevant for visual scenes (for instance Arg1 (utterance) for ``talk'' is not visual), second, auxiliary arguments like direction and manner are not provided (for instance direction and manner for ``look'' are important to describe a scene).
To address this issue, we re-use the MP2D dataset to inform us what arguments are used with the verbs.
For each verb, we choose set of $5$ most frequently used argument role-set and use their union. 
We also remove roles such as TMP (usually referring to words like ``now'', ``then'') since temporal context is implicit in our annotation structure. 
We also removed roles like ADV (adverb) which were too infrequent. 
Finally, we use the following modifier roles: ``Manner'', ``Location'', ``Direction'', ``Purpose'', ``Goal'', but note that ``purpose'' and ``goal'' were restricted to a small number of verbs and hence not considered for evaluation.

We further added the modifier role ``Scene'' which describes ``where'' the event takes place, and only applies to verbs which don't have ``Location''. 
For instance, ``stand'' has the argument role ``location'' which refers to ``where'' the person is standing and doesn't have ``Scene'', whereas ``run'' doesn't contain ``location'' and hence contains ``Scene''. 
In general, ``Scene'' refers to the ``place'' of the event such as ``in an alleyway'' or ``near a beach''.

\textbf{Event Relations.}
We started with the set of three event relations namely: no relation (Events A and B are unrelated), causality (Event B is Caused By Event A \ie B happens directly as a result of A) and contingency based (Event B is Enabled By Event A \ie A doesn't directly cause B but B couldn't have happened without A happening first) on prior work in cross-document event relations \cite{Hong2016BuildingAC}.
However, we found adding an additional case of ``Reaction To'' for causality helpful to distinguish between event relations.
For instance, in the case ``X punches Y'' followed by ``Y falls down'' would be definitely ``B is Caused By A'', however for the case ``X punches Y'' followed by ``Y crouches'' it is unclear if ``B is Caused By A'' since Y makes a voluntary decision to crouch.
As a result, we call this relation ``B is a Reaction To A''.

\subsection{Annotation pipeline}
\label{ss:app_ann_pipe}
With videos, the list of verb-sense and their roles curated, we are now ready to crowd-source annotations on Amazon Mechanical Turk (AMT).

\textbf{Annotation Interface.}
Figure \ref{fig:ann_interface} shows screenshots depicting our annotation interface.
For annotating a given 10 second video, the assigned worker is instructed to first watch the entire 10-second video (Figure \ref{fig:ann_interface} (a)). 
Then for every 2 second interval, the annotator selects a verb corresponding to the most salient event from our curated list of verb-senses using a search-able drop-down menu.
Once the verb is chosen, slots for the corresponding roles are displayed along with an example usage (Figure \ref{fig:ann_interface} (b)). 
The worker fills in the values for each role using free-form text (typically a short phrase). 
When referring to an entity, we instruct the worker to use phrases that uniquely identify the entity in the full 10 second video. 
Furthermore, these phrases can be reused in filling semantic-roles in other events within the video, which provides the co-reference information about the entities \ie co-referenced entities are maintained via exact-string match (Figure \ref{fig:ann_interface} (c)).
Once all verbs and their roles are annotated, we ask the worker to label the relation of Events 1, 2, 4, and 5 with respect to Event 3 (Figure \ref{fig:ann_interface} (d)). 
Note that the order of causality and contingency is different for Events 4,5 compared to Events 1,2 respecting the temporal order.

\textbf{Partitioning into 2-second clips:}
We emphasize that splitting the video into 2-second intervals is strictly a design choice motivated by reduction in annotation cost and consistent quality of annotations.
In an early version of the data collection, we asked annotators to provide ``start'' and ``end'' points for events and allowed overlaps (consistent with other datasets such as ActivityNet Captions\cite{krishna2017dense}).
A close analysis showed that the noise in annotations was tremendous, took significantly longer (roughly 3x) and would lead to a much smaller and lower quality dataset given a budget. 
We thus simplified the task via 2-sec interval annotations and saw large improvements in consensus and speed. 

Clearly, using such a scheme leads to imprecise temporal boundaries for the events. 
Furthermore, it doesn't allow annotating hierarchical actions.
However, we argue that
he \emph{downsides of this design choice are reasonably mitigated} since: 
(a) Longer duration events get annotated via a \emph{repeat} of the same verb across consecutive clips (we see many occurrences in our dataset) \& 
(b) In the presence of multiple verbs in a clip, the most \emph{salient} one gets annotated. 

The 2s duration was chosen after an analysis of ${\sim}50$ videos showed that events typically spanned more than 1s but clips longer than 2s often contained multiple interesting events that we would not want to discard. 
Finally, we note that 2-second duration choice may not be suitable for vastly different domains (e.g. fewer actions and more talking) where 2s may be too dense, and relaxing this to longer clips may be more efficient (annotation cost wise).

\textbf{Event Relation Annotation w.r.t. Middle Event:}
We note there are two alternatives to our proposed annotation strategy for event relation which involves only annotating only all events only with respect to middle event. 
First, exhaustively annotate all event-event relations which would result in $10$ annotations per video. 
Clearly, this is a $2.5 \times$ the annotation (in practice it is even more challenging). 
As a result, we decided to restrict to only one event relation.
Second option is to allow choosing one of the 2-second intervals as the main event and annotating event relations with respect to it. 
In practice, we found the choice of main event to be subjective and inconsistent across annotations. Moreover, choosing the main event could lead to biased event relations (for instance ``Caused By'' relation would be more pronounced).
Thus, we simplified the step by choosing Event 3 spanning from $4$-$6$ seconds as the main event and annotated other events with respect to Event 3. 

\textbf{Worker Qualification and Quality Control.}
To ensure that annotators have understood the task requirements, we put up a qualification task where a worker has to successfully annotate $3$ videos. 
These annotations are manually verified by the first author who then provides feedback on their annotations. 
To filter potential workers, we restrict to more than $95\%$  approval rate and having done at least $500$ tasks.
In total we qualified around $120$ annotators, with at least $60$ workers annotating more than $30$ videos every batch of $2K$ videos.

In addition to manual qualification, we put automated checks one average number of unique verbs provided within a video, and average description lengths. 
We further manually inspect around $3$ random samples from every annotator after every $3K-5K$ videos and provide constant feedback.

\begin{table}[t]
\centering
\begin{tabular}{@{}ccccccc@{}}
\toprule
         & \multicolumn{2}{c}{Acc@1} & \multicolumn{2}{c}{Acc@5} & \multicolumn{2}{c}{Recall@5} \\ 
 & 10 A & \multicolumn{1}{c}{20 A} & 10 A &  \multicolumn{1}{c}{20 A} & 10 A & \multicolumn{1}{c}{20 A} \\ \midrule
Majority & 0.20         & 0.21         & 0.66         & 0.75         & 0.03            & 0.02           \\
Human    & 0.62         & 0.71         & 0.96         & 1.00         & 0.64            & 0.59           \\ \bottomrule
\end{tabular}
\caption{10A and 20A denote 10 and 20 annotations respectively. Majority denotes choosing most frequent verbs for the validation set. }
\label{tab:vb_human_cmp}
\end{table}

\textbf{Annotating Validation and Test Sets.}
We ran a controlled experiments using $100$ videos and annotated $25$ verbs for each event. 
We report the human agreement in Table \ref{tab:vb_human_cmp}.
To compute human agreement score for any event, we use one human annotation (out of 25) as a prediction and the remaining 10 or 20 annotations as ground-truths (denoted by 10A or 20A). 
The final score is the average over all possible prediction/ground-truth partitions.
Essentially, we find that even moving from 10 to 20 annotations, the human agreement improves from $62\%$ to $71\%$ which suggests even at higher number of annotations, we receive verbs which are suggested by a single annotator (and hence no agreement). 
This rules out metrics like accuracy, precision, or F1 scores because they would penalize predictions that may be correct but are not present in a reasonably sized set of ground truth annotations. 
This analysis leads us to the metric Recall@5 which measures if the verbs most agreed upon by humans are indeed recalled by the model in its top-5 predictions.

Furthermore, this prompts us to collect the annotations for validation and test set in two-stages, in the first stage we collect $9$ additional annotations for verb and then in the second-stage $3$ annotations for argument roles and event relations given the verb (we choose the set of verbs chosen by the annotator with the highest agreement, followed by highest number of unique verbs within the video).
We find this two-stage process to be of similar cost of obtaining $5$ independent annotations but with the added advantage of being comparable across annotations.
In total we annotation $3789$ videos for validation and test sets.

\textbf{Reward.}
We set the reward for annotating one 10-second video (for training videos) to $\$0.75$ after estimating the average time of completing an annotation to be around $5$mins. 
This translates to around $\$9$/hour.
Overall, we received generous reviews for the reward on popular turk management website.
For validation and test sets, we set the reward to $\$0.2$ for the first stage (collecting only verbs from $9$ annotators and $\$0.7$ for the second-stage (collecting argument and event relations from $3$ annotators).
As a result, the cost for annotating a single video in the validation and test set turns out to be $\$3.9$ ($0.2 \times 9 + 0.7\times 3$) which is around $5.2\times$ the cost of annotating a single training video.
Total cost for the process comes around $\$36.7K$ (note: this doesn't account for pilot experiments, qualifications, and discarded annotations due to human errors).

\textbf{Collection Timeline.}
Collecting the entire training set was done over a period of about 1.2 months, and an additional 1 month for collecting the validation and test sets.

\section{Additional Dataset Statistics}

In this section we report additional dataset statistics not included in Section \ref{sec:data_stats} due to space constraints.
\begin{table}[t]
\centering
\resizebox{\linewidth}{!}{
\begin{tabular}{@{}cccccc@{}}
\toprule
          & Total & Caused By & Reaction To & Enabled By & No Relation \\ \midrule
Train Set & 94016 & 16.94     & 24.05       & 33.76      & 25.25       \\ \midrule
Val Set   & 5304           & 20.99     & 20.29       & 33.82      & 24.88       \\
Val Set*  & 4089 (77.09\%) & 15.3      & 18.95       & 39.05      & 26.66       \\ \midrule
Test Set  & 6392           & 20.19     & 34.88       & 24.44      & 20.4        \\
Test Set* & 4851 (75.89\%) & 13.39     & 19.04       & 40.9       & 26.5        \\ \bottomrule
\end{tabular}
}
\caption{The distribution of Event Relations before and after filtering by taking consensus of at least two workers \ie we consider only those instances where two workers agree on the event relation when given the verb.}
\label{tab:evrel_hum_agree}
\end{table}

In Table \ref{tab:evrel_hum_agree} we report the distributions of Event Relations before and after filtering for validation and test sets.
For filtering, we use consensus of two workers \ie at least two workers agree on the argument relation which we use as the ground-truth.
We largely find that the consensus on Caused By and Reaction To is low, but Enabled By and No Relations are higher.

\begin{figure*}
    \centering
    \includegraphics[width=\linewidth]{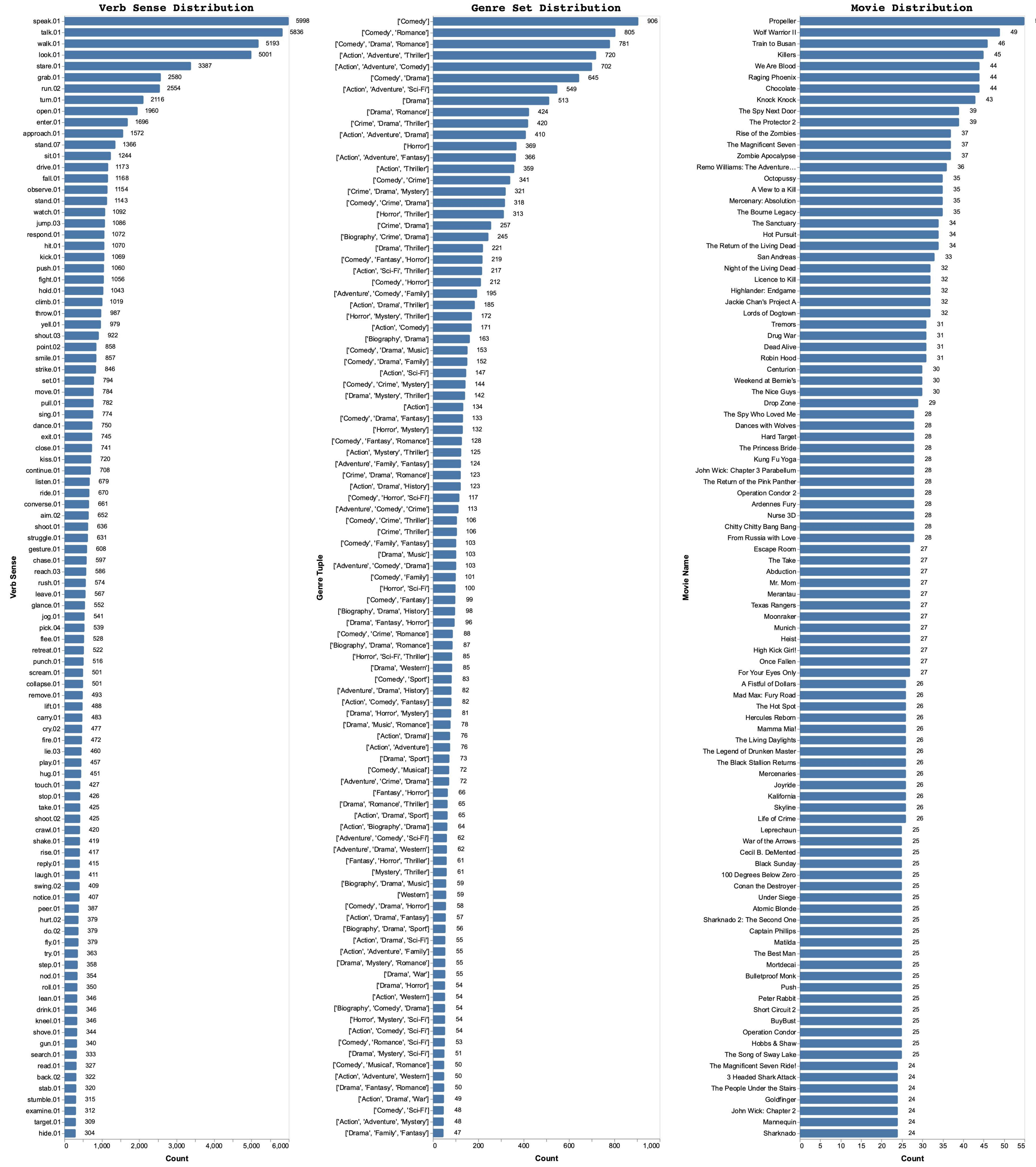}
    \caption{Distribution of $100$ most frequent verbs (a), genre tuples (b), and movies (c). Note that for (a), the count represents the number of events belonging to the particular verb, whereas for (b), (c) it represents the number of video segments belonging to a particular genre or movie.}
    \label{fig:vbgenmov_dist}
\end{figure*}
\begin{figure*}
    \centering
    \includegraphics[width=0.8\linewidth]{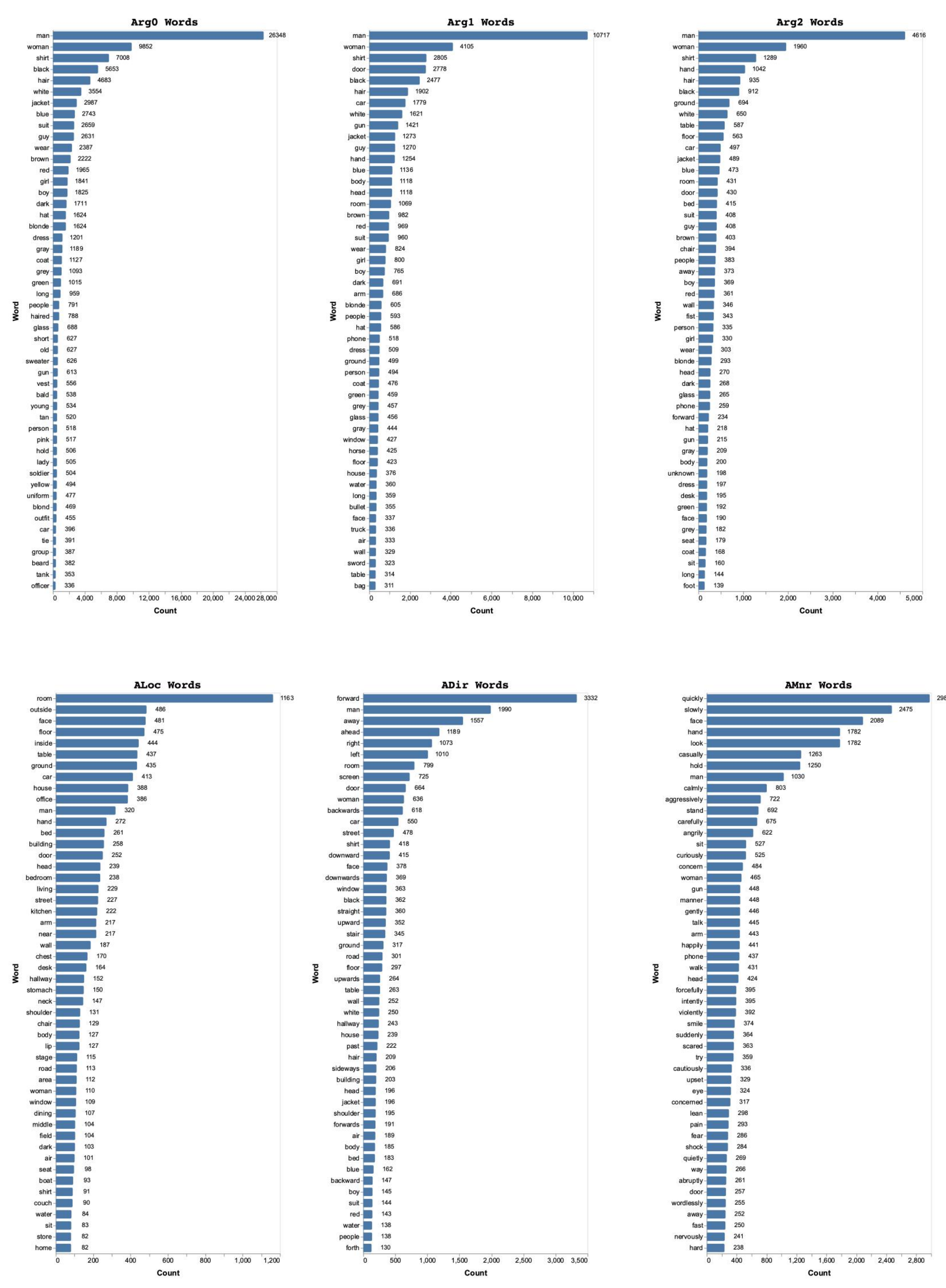}
    \caption{$50$ Most frequent words (after removing stop-words) for Arg0, Arg1, Arg2, ALoc (location), ADir (direction ) and AMnr(Manner).}
    \label{fig:agword_dist}
\end{figure*}

Next, we plot the distributions for the $100$ most frequent verbs, genres and chosen movies in Figure \ref{fig:vbgenmov_dist}.
For verbs and genres we find Zipf's law in action. 
For verbs, we find most common verbs such as ``talk'', ``speak'', ``walk'', ``look'' which are also part of frequent atomic actions despite explicitly not scoring them. 
This is an inherent effect due to the movie domain where dialogue is a large focus.
For genres we find that ``Comedy'', ``Drama'', ``Action'', ``Romance'' are the most frequent which tend to have more movements than ``Mystery'', ``Thriller'' which have less movements on actors with often extended still-frames.

In Figure \ref{fig:agword_dist} we plot the top $50$ most frequent words within the argument (after removing stop-words). 
We find ``man'', ``woman'' are the most frequent word in all of Arg0, Arg1, Arg2 which is not surprising since the movies are human-centric.
We note the over-abundance of ``man'' compared to ``woman'' is an amplification of the biases present in the movie.
Interestingly, the distribution is less skewed for Location, Direction, and Manner
\section{Implementation Details}
We detail some of the implementation details for our models.
All implementations are coded in PyTorch \cite{NEURIPS2019_9015_Pytorch}.
Unless otherwise mentioned we use Adam \cite{Kingma2015AdamAM} optimizer with learning rate of $1e^{-4}$. 

\subsection{Verb Prediction Models}
All our implementations for verb prediction models such as I3D\cite{Carreira2017QuoVA}, Slow-only and SlowFast networks \cite{Feichtenhofer2019SlowFastNF} is based on the excellent repository SlowFast \cite{fan2020pyslowfast}.
We use the checkpoints from the repository for kinetics pre-trained models.
All models are trained with a batch size of $8$ for $10$ epochs, and the model with best recall@5 is chosen for testing.
For classification, we use a set of $1560$ verbs composed two MLP projections (first projects to half the input dimension, the second to $1560$ verbs)  separated with a ReLU activation.
For inference, we choose the top-5 scoring verbs.
Training requires considerable GPU space, and on $8$ TITAN GPUs, with batch size of $8$ each epoch takes around $1$ hour, with total being $10$ hours.

\subsection{Argument Prediction Models}
We extract the features from underlying base networks which is $2048$ and $2304$ for I3D and SlowFast respectively.
For transformers, we use the implementation provided in Fairseq library \cite{ott2019fairseq} (\footnote{\url{https://github.com/pytorch/fairseq/}} and for GPT2 (medium) and Roberta (base) we use the implementation by HuggingFace transformer library \cite{wolf-etal-2020-transformers} \footnote{\url{https://github.com/huggingface/transformers}}.
For tokenization and vocabulary, we utilize Byte-Pair Encoding and add special argument tokens such as $[Arg0]$ to encode the phrases.

For both transformer encoder and decoder we use $3$ layers with $8$ attention heads.
The decoder uses the last encoder layer outputs as encoder attention for subsequent decoding.
For training, we use cross-entropy loss over the predicted sequence.
For sequence generation, we use greedy-decoding with temperature $1.0$ as we didn't find improvements using beam-search or using different temperature.

For training, we used a batch size of $16$ for all models other than GPT2 for which we could only use a batch size of $8$ due to memory restrictions.
Training time for GPT2 is around $10$ hours over $8$ GPUs (recall that GPT2 medium has $24$ transformer layers and $16$ attention heads).
All other models take around $15$ mins per epoch with batch size of $16$ on a single TITAN GPU with total time around $3$ hours for $10$ epochs which we found sufficient for convergence.

For computing natural language generation metrics like ROUGE, CIDEr we use the official MSCOCO Captions implementation \cite{Lin2014MicrosoftCoCo} \footnote{\url{https://github.com/tylin/coco-caption}}.
For co-reference metrics, we use the implementation provided in coval \cite{Moosavi2016WhichCE} \footnote{\url{https://github.com/ns-moosavi/coval}}
\section{Evaluation Metrics}
In this section, we provide details on LEA as well as our proposed LEA-soft. 
We further report additional metrics such BLEU \cite{Papineni2002BleuAM} and METEOR \cite{Banerjee2005METEORAA}, and coreference metrics.
We also report per-argument scores for the baselines.

\subsection{Co-Reference Metrics}
We primarily use the metric LEA \cite{Moosavi2016WhichCE} which is a link-based metrics. 
We also note there exists other metrics such as
MUC~\cite{Vilain1995AMC}, BCUBE~\cite{Bagga1998EntityBasedCC}, CEAFE\cite{luo-2005-coreference}. 
We point the reader to a seminal paper on visualizing these metrics ~\cite{pradhan-etal-2014-scoring} for a brief overview of MUC, BCUBE and CEAFE, and \cite{Moosavi2016WhichCE} for comparison of other metrics with LEA.

\textbf{LEA and LEA-soft}
As noted in the paper \cite{Moosavi2016WhichCE}, LEA computes an importance score and resolution score for each entity given as 
\begin{align}
    \frac{\sum_{e_i \in E} imp(e_i) \times res(e_i) } {\sum_{e_i \in E} imp(e_i)}
\end{align}

The final score is the F1-measure computed based on recall (entities are ground-truths) and precision (entities are predictions).
As noted earlier, LEA doesn't consider if the proposed entity by itself is correct and thus even incorrect entity predictions could lead high co-reference score as long as the co-referencing is correct.
We address this using LEA-soft which additionally weights the importance of each entity during precision computation with the sum of cider scores in the numerator and len of cider scores in the denominator.

As a result, we have
\begin{align}
    Prec_{LEA} &= \frac{\sum_{e_i \in E} imp(e_i) \times res(e_i) } {\sum_{e_i \in E} imp(e_i)} \\
    Prec_{LEA-soft} &= \frac{\sum_{e_i \in E}(\sum_{e_i} C(e_i)) \times  imp(e_i) \times res(e_i) } {\sum_{e_i \in E} |e_i| \times imp(e_i)}
\end{align}
where $C(e_i)$ denotes the cider score for the $i^{th}$ entity. 
We keep the recall computation unchanged and use the modified precision to compute the final F1-Score for LEA-soft.
Since we have multiple ground-truth reference, we compute the F1-score for each ground-truth reference individually and average over the $3$ ground-truths.

\subsection{Evaluation of Arguments}

\begin{table}[t]
\centering
\resizebox{\linewidth}{!}{
\begin{tabular}{@{}c|cccccccc}
\toprule
      & cider & Arg0 & Arg1 & Arg2 & ALoc & AScn & ADir & AMnr  \\ \midrule
GPT2  & 0.39  & 0.40 & 0.39 & 0.45 & 0.43 & 0.22 & 0.37 & 0.15  \\
Human & 0.70  & 0.73 & 0.74 & 0.73 & 0.90 & 0.96 & 0.40 & 0.15  \\ \bottomrule
\end{tabular}
}
\caption{CIDEr score for all collected Arguments with $5$ annotations on $100$ videos.}
\label{tab:vbarg_human_cmp}
\end{table}

We examine the cider scores for different arguments over a set of $100$ videos (same used for verb prediction results).
To compare semantic role values, which are free-form text phrases, we compute CIDEr metric treating one of the chosen annotations as a hypothesis and remaining annotations as references for each argument.
Table~\ref{tab:vbarg_human_cmp} compares CIDEr scores for all semantic roles and scores by argument type for a GPT2 based language only baseline that generates the sequence of roles and values given the verb for an event. 
We find that human-agreement is high for all arguments except direction (ADir) and manner (AMnr). 
For both ``direction" (ADir) and ``manner" (AMnr), we find that both language-only baseline and human agreements are poor. 
On further inspection, we find that the argument ``manner'' describes ``how'' the event took place is open to subjective interpretation, and the argument ``direction'' has a wide range of correct values (e.g. for ``walk'' directions ``forward'', ``down the path'', and ``through the trees'') may all be correct. 
For a reliable evaluation, we evaluate argument prediction performance only on arguments that achieved high human-agreement \ie Arg0, Arg1, Arg2, ALoc, and AScn, and leave the evaluation of Direction and Manner for future work.\\

\subsection{All Metrics}
\begin{table*}[t]
\centering
\begin{tabular}{@{}c|cc|cc|cc|c@{}}
\toprule
Model     & GPT2   & TxDec  & Vid TxDec & Vid TxEncDec & Vid TxDec & Vid TxEncDec & Human \\ 
Vis Feats & \xmark & \xmark & SlowFast  & SlowFast     & I3D       & I3D          &       \\ 
\midrule
B@1       & 40.91  & 42.79  & 43.45     & 44.65        & 41.69     & 45.3         & 43.56 \\
B@1-Vb    & 38.08  & 41.02  & 39.59     & 41.98        & 38.96     & 40.54        & 39.93 \\
B@1-Arg   & 40.91  & 42.62  & 42.89     & 44.49        & 40.18     & 44.6         & 41.69 \\
B@1-Arg0  & 44.67  & 46.32  & 48.26     & 48.14        & 49.58     & 49.36        & 49.71 \\
B@1-Arg1  & 31.88  & 31.69  & 32.81     & 34.72        & 34.76     & 36.17        & 40.61 \\
B@1-Arg2  & 34.13  & 36.3   & 34.93     & 35.86        & 35.17     & 37.36        & 39.87 \\
B@1-ALoc  & 46.88  & 48.07  & 48.97     & 51.39        & 42.73     & 49.37        & 38.7  \\
B@1-AScn  & 46.99  & 50.74  & 49.48     & 52.33        & 38.66     & 50.71        & 39.56 \\
\midrule
B@2       & 27.66  & 28.8   & 29.87     & 30.86        & 28.47     & 30.73        & 29.89 \\
B@2-Vb    & 23.92  & 26.52  & 25.73     & 27.54        & 25        & 25.39        & 25.14 \\
B@2-Arg   & 27.63  & 28.4   & 29.19     & 30.61        & 26.82     & 30.06        & 28.37 \\
B@2-Arg0  & 31.06  & 32.07  & 34.09     & 33.78        & 35.33     & 34.03        & 34.74 \\
B@2-Arg1  & 19.53  & 19.87  & 20.25     & 22.39        & 22.3      & 22.6         & 26.72 \\
B@2-Arg2  & 22.1   & 23.52  & 22.22     & 23.46        & 21.81     & 24           & 26.76 \\
B@2-ALoc  & 32.92  & 32.24  & 34.19     & 35.98        & 28.58     & 34.04        & 27.06 \\
B@2-AScn  & 32.55  & 34.29  & 35.21     & 37.42        & 26.06     & 35.61        & 26.59 \\
\midrule
M         & 16.99  & 17.51  & 17.28     & 18.26        & 17.68     & 18.32        & 22.24 \\
M-Vb      & 15.33  & 16.4   & 15.8      & 17.14        & 16.39     & 16.77        & 22.08 \\
M-Arg     & 15.88  & 16.03  & 16.2      & 17.23        & 15.93     & 16.95        & 21.02 \\
M-Arg0    & 21.12  & 21.97  & 20.99     & 21.46        & 22.23     & 22.05        & 25.21 \\
M-Arg1    & 15.49  & 14.81  & 13.94     & 16.14        & 15.93     & 16.16        & 22.22 \\
M-Arg2    & 14.99  & 16.27  & 15.21     & 15.65        & 14.76     & 14.85        & 20.75 \\
M-ALoc    & 15.21  & 13     & 15.03     & 16.26        & 12.17     & 15.19        & 17.88 \\
M-AScn    & 12.59  & 14.11  & 15.85     & 16.63        & 14.54     & 16.51        & 19.02 \\
\midrule
R         & 40.08  & 41.19  & 40.61     & 42.66        & 40.67     & 42.41        & 39.77 \\
R-Vb      & 37.07  & 37.89  & 36.89     & 39.18        & 36.38     & 38.14        & 39.16 \\
R-Arg     & 39.62  & 40.47  & 39.58     & 41.96        & 38.56     & 41.39        & 38.43 \\
R-Arg0    & 44.77  & 46.7   & 46.78     & 47.36        & 48.65     & 47.71        & 45.84 \\
R-Arg1    & 34.25  & 33.24  & 32.83     & 35.7         & 34.66     & 36.65        & 40.23 \\
R-Arg2    & 33.72  & 36.14  & 34.12     & 35.13        & 34.71     & 35.85        & 36.43 \\
R-ALoc    & 42.87  & 41.41  & 39.82     & 44.6         & 32.22     & 41.49        & 34.38 \\
R-AScn    & 42.46  & 44.84  & 44.33     & 46.99        & 42.55     & 45.26        & 35.25 \\
\midrule
C         & 34.67  & 35.68  & 44.78     & 45.52        & 47.14     & 47.06        & 84.85 \\
C-Vb      & 42.97  & 47.5   & 49.97     & 55.47        & 51.61     & 51.67        & 91.7  \\
C-Arg     & 34.45  & 32.15  & 41.24     & 42.82        & 41.29     & 42.76        & 80.15 \\
C-Arg0    & 28.33  & 32.1   & 41.64     & 34.6         & 48.99     & 39.42        & 88.24 \\
C-Arg1    & 38.58  & 38.47  & 41.42     & 45.47        & 45.42     & 47.06        & 83.37 \\
C-Arg2    & 36.82  & 40.51  & 42.28     & 41.02        & 40.19     & 44.52        & 74.82 \\
C-ALoc    & 47.77  & 27.05  & 43.01     & 46.97        & 33.75     & 39.75        & 76.72 \\
C-AScn    & 20.73  & 22.62  & 37.86     & 46.05        & 38.11     & 43.03        & 77.62 \\
\midrule
MUC       & 59.13  & 64.54  & 45.59     & 65.48        & 46.01     & 61.57        & 80.75 \\
BCUBE     & 73.53  & 74.43  & 69.39     & 72.97        & 68.74     & 73.34        & 86.32 \\
CEAFE     & 61.75  & 63.84  & 57.26     & 59.7         & 56.2      & 61.16        & 77.8  \\
LEA       & 48.08  & 51.76  & 37.88     & 50.48        & 37.89     & 48.92        & 72.1  \\
LEA Soft  & 28.1   & 28.6   & 28.69     & 31.99        & 30.38     & 33.58        & 70.33 \\
\bottomrule
\end{tabular}
\caption{{\footnotesize Semantic Role Prediction on Validation Set. B@1: Bleu-1, B@2: Bleu-2, M: METEOR, R: ROUGE-L, C: CIDEr, Metric-Vb: Macro Averaged over Verbs, Metric-Arg: Macro Averaged over arguments, Metric-Argi: Metric computed only for the particular argument.}}
\label{tab:vb_arg_all_mets_val}
\end{table*}

\begin{table*}[t]
\centering
\begin{tabular}{@{}c|cc|cc|cc|c@{}}
\toprule
Model     & GPT2   & TxDec  & Vid TxDec & Vid TxEncDec & Vid TxDec & Vid TxEncDec & Human \\
Vis Feats & \xmark & \xmark & SlowFast  & SlowFast     & I3D       & I3D          &       \\ \midrule
B@1       & 41.89  & 42.9   & 43.4      & 45.36        & 43.69     & 45.56        & 43.46 \\
B@1-Vb    & 38.41  & 39.4   & 39.28     & 41.03        & 39.43     & 40.52        & 39.73 \\
B@1-Arg   & 41.9   & 42.56  & 42.84     & 45.25        & 42.04     & 44.83        & 41.47 \\
B@1-Arg0  & 45.65  & 46.06  & 47.56     & 48.92        & 48.96     & 49.75        & 48.2  \\
B@1-Arg1  & 32.17  & 31.53  & 33.15     & 34.46        & 33.93     & 35.42        & 41.06 \\
B@1-Arg2  & 35.02  & 37.34  & 34.85     & 36.69        & 36.32     & 38.55        & 39.69 \\
B@1-ALoc  & 48.7   & 46.53  & 48.74     & 52.95        & 43.91     & 49.18        & 36.74 \\
B@1-AScn  & 47.94  & 51.34  & 49.88     & 53.23        & 47.07     & 51.25        & 41.65 \\
\midrule
B@2       & 28.43  & 29.15  & 30.08     & 31.64        & 30.34     & 31.34        & 29.43 \\
B@2-Vb    & 24.25  & 25.49  & 25.83     & 26.9         & 25.45     & 26.22        & 24.37 \\
B@2-Arg   & 28.41  & 28.7   & 29.42     & 31.56        & 28.79     & 30.59        & 27.95 \\
B@2-Arg0  & 31.69  & 31.92  & 33.56     & 34.33        & 34.84     & 34.76        & 32.99 \\
B@2-Arg1  & 19.8   & 19.88  & 20.98     & 22.69        & 22.3      & 22.46        & 26.88 \\
B@2-Arg2  & 22.43  & 24.39  & 22.36     & 24.15        & 23.05     & 24.81        & 26.27 \\
B@2-ALoc  & 34.36  & 31.63  & 34.18     & 37.95        & 30.69     & 34.32        & 25.66 \\
B@2-AScn  & 33.76  & 35.67  & 36.03     & 38.66        & 33.05     & 36.62        & 27.93 \\
\midrule
M         & 17.74  & 17.67  & 17.45     & 18.83        & 18.22     & 18.7         & 21.86 \\
M-Vb      & 15.8   & 15.84  & 15.72     & 17.02        & 16.92     & 16.83        & 22.44 \\
M-Arg     & 16.63  & 16.21  & 16.46     & 17.9         & 16.63     & 17.44        & 20.55 \\
M-Arg0    & 21.82  & 21.83  & 20.72     & 21.96        & 22.2      & 22.23        & 24.61 \\
M-Arg1    & 15.99  & 14.97  & 14.39     & 16.31        & 16.28     & 16.53        & 21.55 \\
M-Arg2    & 15.39  & 16.63  & 15.15     & 16.22        & 15.34     & 15.41        & 20.11 \\
M-ALoc    & 16.41  & 12.96  & 15.76     & 17.63        & 13.59     & 16.2         & 16.89 \\
M-AScn    & 13.55  & 14.68  & 16.3      & 17.36        & 15.74     & 16.82        & 19.58 \\
\midrule
R         & 41.33  & 41.45  & 41.12     & 43.46        & 41.5      & 42.96        & 40.04 \\
R-Vb      & 37.71  & 36.96  & 36.66     & 38.6         & 36.69     & 37.72        & 39.24 \\
R-Arg     & 40.91  & 40.65  & 40.14     & 42.88        & 39.68     & 42.04        & 38.55 \\
R-Arg0    & 45.89  & 46.6   & 46.75     & 48.22        & 48.69     & 48.3         & 45.5  \\
R-Arg1    & 35.13  & 33.05  & 33.35     & 35.67        & 34.9      & 36.34        & 40.03 \\
R-Arg2    & 34.13  & 36.83  & 33.77     & 35.26        & 35.58     & 36.49        & 37.29 \\
R-ALoc    & 45.33  & 40.96  & 41.53     & 47.17        & 35.1      & 43.06        & 32.94 \\
R-AScn    & 44.04  & 45.82  & 45.31     & 48.08        & 44.14     & 46.04        & 36.97 \\
\midrule
C         & 36.48  & 35.34  & 44.95     & 47.25        & 47.9      & 48.51        & 83.68 \\
C-Vb      & 44.27  & 44.44  & 49.46     & 52.92        & 51.29     & 53.88        & 87.78 \\
C-Arg     & 36.51  & 32.06  & 41.98     & 45.48        & 43.62     & 44.53        & 79.29 \\
C-Arg0    & 26.17  & 27.83  & 36.84     & 33.51        & 41.89     & 38.64        & 81.62 \\
C-Arg1    & 39.08  & 37.99  & 42.93     & 43.79        & 46.53     & 46.47        & 81.47 \\
C-Arg2    & 35.36  & 41.93  & 39.16     & 39.48        & 41.66     & 43.84        & 73.21 \\
C-ALoc    & 55.05  & 25.83  & 48.3      & 58.38        & 43.83     & 45.15        & 77.38 \\
C-AScn    & 26.9   & 26.71  & 42.65     & 52.22        & 44.18     & 48.57        & 82.77 \\
\midrule
MUC       & 60.51  & 65.42  & 47.51     & 65.91        & 47.63     & 62.62        & 80.8  \\
BCUBE     & 74.21  & 74.76  & 69.84     & 72.95        & 69.2      & 73.6         & 86.26 \\
CEAFE     & 62.19  & 63.85  & 57.33     & 59.57        & 56.65     & 61.41        & 77.38 \\
LEA       & 49.38  & 52.46  & 38.91     & 50.88        & 38.77     & 49.61        & 71.77 \\
LEA Soft  & 30.24  & 29.18  & 30.21     & 33.5         & 31.73     & 35.46        & 70.6 
\\ \bottomrule
\end{tabular}
\caption{{\footnotesize Semantic Role Prediction on Test Set. B@1: Bleu-1, B@2: Bleu-2, M: METEOR, R: ROUGE-L, C: CIDEr, Metric-Vb: Macro Averaged over Verbs, Metric-Arg: Macro Averaged over arguments, Metric-Argi: Metric computed only for the particular argument.}}

\label{tab:vb_arg_all_mets_test}
\end{table*}
We report BLEU@1, BLUE@2, METEOR, ROUGE, and CIDEr for both val (Table \ref{tab:vb_arg_all_mets_val}) and test set (Table \ref{tab:vb_arg_all_mets_test}).
For each metric we further report macro-averaged scores across verbs and arguments, and report per argument scores.
Note that only CIDEr is able to take advantage of the macro-averaged scores due to its inverse document frequency re-weighting.
Finally, we report the co-reference metrics MUC, BCUBE, CEAFE , LEA and our proposed metric LEA-Soft.

\section{\dsn{} DataSheet}
The seminal work datasheets for datasets \cite{Gebru2018DatasheetsFD} outlines a list of questions to encourage  transparency, accountability and mitigate unwanted biases.
Here, we provide a datasheet for \dsn{} closely following the guidelines in prior work.
For simplicity and readability, we paste the questions verbatim.

\subsection{Motivation}
\begin{itemize}
    \itemsep0em
    \item \textbf{For what purpose was the dataset created?} 
    The main motivation to create the dataset is to bridge the research gap between learning atomic actions and generating holistic captions. 
    In particular, the dataset opens path for the task of \tkfull{} which in addition to action-recognition, emphasizes how various objects interact within an action, how various objects interact over time-period across multiple actions, co-referencing of these objects over time, and how various actions affect each other.
    
    \item \textbf{Who created the dataset (e.g., which team, research group) and on
behalf of which entity (e.g., company, institution, organization)?} 
The dataset is created by the authors who belong to PRIOR Team at AI2. The first-author (Arka Sadhu) was a summer intern in the PRIOR Team.
    \item \textbf{Who funded the creation of the dataset?}
    PRIOR Team at AI2 funded the creation of the dataset.
\end{itemize}

\subsection{Composition}
\begin{itemize}
    \itemsep0em
    \item \textbf{What do the instances that comprise the dataset represent (e.g.,
documents, photos, people, countries)?}  
Each instance consists of a 10-second video obtained from a movie-clip available on YouTube. 
These are usually human-centric and hence primarily contain videos of people interacting in diverse and complex situations. 

    \item \textbf{How many instances are there in total (of each type, if appropriate)?}
    In total there are $29.2K$ instances distributed across training ($23.62K$), validation ($1.32K$) and testing ($4.1K$). Individual test sets are available for each of the sub-task. 
    \item \textbf{Does the dataset contain all possible instances or is it a sample
(not necessarily random) of instances from a larger set?} This question doesn't pertain to our dataset.
    \item \textbf{What data does each instance consist of?} 
    Each instance is a $10$-second video (mp4 video) available from YouTube.
    \item \textbf{Is there a label or target associated with each instance?} 
    Each instance ($10$ second video) is annotated at $2$-second intervals with a verb describing the event, corresponding argument roles for the verb co-referenced across the video, and event relations across the various verbs with respect to the middle event (Event 3 spanning from 4-6 seconds).
    \item \textbf{Is any information missing from individual instances?} No, every instance has the same annotations. 
    \item \textbf{Are relationships between individual instances made explicit
(e.g., users’ movie ratings, social network links)?}
    We provide information about which instances are derived from the same $2-3$ minutes YouTube video as well as the underlying movie (this information is obtained from Condensed-Movies \cite{bain2020condensed} dataset).
    However, this information is not used for any of the task in the dataset except for splitting the videos in train, validation and test sets.
    \item \textbf{Are there recommended data splits (e.g., training, development/validation, testing)?} Yes, we provide training, validation and test sets by splitting the overall set in $80:5:15$ ratio randomly based on the movie names. We also ensure (qualitatively) that the normalized distributions of verbs, and genres are same across the splits.
    \item \textbf{Are there any errors, sources of noise, or redundancies in the dataset?} The main sources of errors would be the annotations themselves, however, we have made extended efforts from automatic to manual checks to remove such errors and provided constant feedback. Some redundancy may occur due to oversampling of dialogues in movies which are described with the verb ``talk''. Some redundancy may also occur due to use of closely related verbs such as ``run'' and ``jog''.
    \item \textbf{Is the dataset self-contained, or does it link to or otherwise rely on
external resources (e.g., websites, tweets, other datasets)?}
    Yes, the dataset provides links to YouTube videos. Since the videos are provided by a licensed channel, we expect the videos to have high online longevity.
    \item \textbf{Does the dataset contain data that might be considered confidential (e.g., data that is protected by legal privilege or by doctor patient confidentiality, data that includes the content of individuals’ non-public communications)?} No, our dataset is derived from movies publicly available on youtube.
    \item \textbf{Does the dataset contain data that, if viewed directly, might be offensive, insulting, threatening, or might otherwise cause anxiety?} Some of the videos obtained from action, crime or horror movies may be sensitive to some viewers when viewed directly. 
    Some videos may also contain violence and gore, and we suggest user discretion in viewing the videos.
    
\end{itemize}

\subsection{Collection Process}
\begin{itemize}
    \itemsep0em
    \item \textbf{How was the data associated with each instance acquired?}
    The data was directly observable in the form of embedded youtube videos.
    \item \textbf{What mechanisms or procedures were used to collect the data
(e.g., hardware apparatus or sensor, manual human curation, software program, software API)? } We used Amazon Mechanical Turk to collect the data with a custom annotation interface. We validated them by small scale user study and taking feedbacks during worker qualification.
    \item \textbf{If the dataset is a sample from a larger set, what was the sampling
strategy (e.g., deterministic, probabilistic with specific sampling
probabilities)?} We sampled videos which had more verbs within their duration.
    \item \textbf{Who was involved in the data collection process (e.g., students,
crowdworkers, contractors) and how were they compensated (e.g.,
how much were crowdworkers paid)?} Crowd-Workers were involved in the process. They were paid $\$0.75$ for training videos and $\$0.2$ for verb annotation and $\$0.7$ for argument and event relation for videos in validation and test splits. On average it is around $\$9-\$12$ per hour above the minimum wage.
On popular websites, our pay was noted to be generous.
    \item  \textbf{Over what timeframe was the data collected?} The data was collected over $2.2$ months with initial $1.2$ months for training set and rest for validation and testing.
    \item \textbf{Were any ethical review processes conducted (e.g., by an institutional review board)?} No, there was no ethical review process.
    
\end{itemize}

\subsection{Preprocessing/cleaning/labeling}

\begin{itemize}
    \itemsep0em
    \item \textbf{Was any preprocessing/cleaning/labeling of the data done (e.g.,
discretization or bucketing, tokenization, part-of-speech tagging,
SIFT feature extraction, removal of instances, processing of missing values)? } Only,  exact string match was performed to obtain co-referenced entities. We used spacy \cite{spacy2} to compute dataset statistics such as noun-diversity but it is not used over the collected data for down-stream tasks.
    \item \textbf{Was the “raw” data saved in addition to the preprocessed/cleaned/labeled
data (e.g., to support unanticipated future uses)?} In our case, raw data is same as cleaned data.
\end{itemize}

\subsection{Uses}
\begin{itemize}
    \itemsep0em
    \item \textbf{Has the dataset been used for any tasks already? }
    We have used the data to show its usefulness for our proposed task \tkfull{}
    \item \textbf{Is there a repository that links to any or all papers or systems that
use the dataset?} 
    Updated information about the dataset can be found on \vweb{}.
    \item \textbf{What (other) tasks could the dataset be used for?} We believe the dataset could be re-purposed for many down-stream video understanding tasks such as video retrieval, video question answering, action forecasting, long-term reasoning.
    \item \textbf{Are there tasks for which the dataset should not be used?} The data is obtained from movies and exhibits certain stereotypes which donot hold true in real world. It also contains highly unlikely action sequences (such as a ``man flying''), and thus it shouldn't be used for real-world cases and strictly used as a video understanding benchmark.
\end{itemize}

\subsection{Distribution}
\begin{itemize}
    \itemsep0em
    \item \textbf{Will the dataset be distributed to third parties outside of the entity (e.g., company, institution, organization) on behalf of which
the dataset was created?} The dataset is publicly available at \vweb{}.
    \item \textbf{How will the dataset will be distributed (e.g., tarball on website,
API, GitHub)?} The dataset is available through our website and github. The dataset is stored on Amazon S3 buckets.
\end{itemize}

\newpage
{\small
\bibliographystyle{ieee_fullname}
\bibliography{egbib,ref}
}

\end{document}